\documentclass[10pt,journal,compsoc]{IEEEtran}
\ifCLASSOPTIONcompsoc
\usepackage[nocompress]{cite}
\else
\usepackage{cite}
\fi
\ifCLASSINFOpdf
\else
\fi

\usepackage{url}
\usepackage{hyperref}
\usepackage{graphicx}
\usepackage{caption}
\usepackage{subcaption}
\usepackage{bm}
\usepackage{amssymb}
\usepackage{amsmath}
\usepackage{booktabs}
\usepackage{diagbox}
\usepackage{multirow}
\usepackage{textcomp}
\usepackage{algorithm}
\usepackage{algorithmic}
\usepackage{color}
\usepackage{amsthm}
\newtheorem{lemma}{Lemma}

\begin{document}
	
\title{Cross-Layer Distillation with Semantic Calibration}
%
%

\author{Defang Chen, Jian-Ping Mei, Yuan Zhang, Can Wang, Yan Feng, Chun Chen
\IEEEcompsocitemizethanks{
\IEEEcompsocthanksitem Defang Chen, Yuan Zhang, Can Wang, Yan Feng and Chun Chen are with the College of Computer Science, Zhejiang University, Hangzhou 310027, China.
\protect\\
		E-mail: \{defchern, yuan\_zhang, wcan, fengyan, chenc\}@zju.edu.cn. 
		\IEEEcompsocthanksitem Jian-Ping Mei is with the College of Computer Science, Zhejiang University of Technology, Hangzhou, 310027, China. \protect\\
		E-mail: jpmei@zjut.edu.cn.}
	\thanks{(Corresponding author: Can Wang.)}
}

%
%

\markboth{Journal of \LaTeX\ Class Files,~Vol.~xx, No.~x, August~2021}%
{Chen \MakeLowercase{\textit{et al.}}: Cross-Layer Distillation with Semantic Calibration}
%


\IEEEtitleabstractindextext{%
\begin{abstract}
Knowledge distillation is a technique to enhance the generalization ability of a student model by exploiting outputs from a teacher model. 
Recently, feature-map based variants explore knowledge transfer between manually assigned teacher-student pairs in intermediate layers for further improvement. 
However, layer semantics may vary in different neural networks and semantic mismatch in manual layer associations will lead to performance degeneration due to negative regularization. To address this issue, we propose Semantic Calibration for cross-layer Knowledge Distillation (SemCKD), which automatically assigns proper target layers of the teacher model for each student layer with an attention mechanism. 
With a learned attention distribution, each student layer distills knowledge contained in multiple teacher layers rather than a specific intermediate layer for appropriate cross-layer supervision. 
We further provide theoretical analysis of the association weights and conduct extensive experiments to demonstrate the effectiveness of our approach. Code is avaliable at \url{https://github.com/DefangChen/SemCKD}.
\end{abstract}
	
\begin{IEEEkeywords}
	Knowledge Distillation, Semantic Calibration, Cross-Layer Distillation, Attention Mechanism.
\end{IEEEkeywords}}

\maketitle

\IEEEdisplaynontitleabstractindextext

%
\IEEEpeerreviewmaketitle

\IEEEraisesectionheading{\section{Introduction}\label{sec:introduction}}
\IEEEPARstart{H}{ow} to improve the generalization ability and compactness of deep neural networks keeps attracting widespread attention among a large number of researchers \cite{bengio2013representation,han2016deep,zhang2017understanding}. In the past few years, a lot of techniques have sprung up to deal with these two issues, making the deep models become more powerful \cite{he2015delving,srivastava2014dropout,loffe2015batch,zhang2018mixup} or more portable \cite{denil2013predicting,hinton2015distilling,han2015learning,polino2018model}. Among these techniques, knowledge distillation (KD) provides a promising solution to simultaneously achieve better model performance as well as maintain simplicity \cite{hinton2015distilling}.

The insightful observation that the generalization ability of a lightweight model can be improved by training to match the prediction of a powerful model stems from some pioneering works on model compression \cite{bucilua2006model,ba2014deep}. Recently this idea is popularized by knowledge distillation in which the powerful model named as \textit{teacher} is pre-trained and its outputs are exploited to improve the performance of the lightweight model named as \textit{student} \cite{hinton2015distilling}. Instead of using discrete labels, the student model employs class predictions from the teacher model as an effective regularization to avoid trapping in over-confident solutions \cite{rafael2019when,yuan2020revisiting}. 
A major drawback in the vanilla KD is that the knowledge learned by a classification model is represented only by the prediction of its final layer \cite{hinton2015distilling}. Although the relative probabilities assigned to different classes provide an intuitive understanding about model generalization, knowledge transfer in such a highly abstract form overlooks a wealth of information contained in intermediate layers. 
To further boost the effectiveness of distillation, recent 
works explore various representations of feature maps to capture the enriched knowledge
\cite{romero2015fitnet,zagoruyko2017paying,tung2019similarity,ahn2019variational,passalis2020heterogeneous,yue2020mgd}. An interpretation for the success of feature-map based knowledge distillation is that the multi-layer feature representations respect hierarchical concept learning process and will entail reasonable inductive bias \cite{bengio2013representation,farabet2013learning}. 


However, intermediate layers in teacher and student models tend to have different levels of abstraction \cite{tung2019similarity,kornblith2019similarity,passalis2020heterogeneous}. A particular challenge in feature-map distillation is thus to ensure appropriate layer associations for maximum performance. 
But existing efforts still enable knowledge transfer based on hand-crafted layer assignments \cite{romero2015fitnet,tung2019similarity,zagoruyko2017paying,ahn2019variational,passalis2020heterogeneous,yue2020mgd}. Such strategies may cause semantic mismatch in certain teacher-student layer pairs, leading to negative regularization in the student model training and deterioration of its performance. As we have no prior knowledge of intermediate layer semantics, layer association becomes a non-trivial problem. Systematic approaches need to be developed for more effective and flexible knowledge transfer using feature maps.

In this paper, we propose \textbf{Sem}antic \textbf{C}alibration for cross-layer \textbf{K}nowledge \textbf{D}istillation (SemCKD) to exploit intermediate knowledge by keeping the transfer in a matched semantic level. An attention mechanism is applied to automatically learn soft layer association, which effectively binds a student layer with multiple semantic-related target layers rather than a fixed one in the teacher model. 
To align the spatial dimensions of feature maps in each layer pair for calculating the total loss, feature maps of the student layers are projected to the same dimensions as those in the associated target layers. By taking advantage of semantic calibration and feature-map transfer across multiple layers, the student model can be effectively optimized with more appropriate guidance. 
The overall contributions of this paper are summarized as follows: 
\begin{itemize}
	\item We propose a novel technique to significantly improve the effectiveness of feature-map based knowledge distillation. Our approach is readily applicable to heterogeneous settings where different architectures are used for teacher and student models.
	\item Learning soft layer association by our proposed attention mechanism can effectively alleviate semantic mismatch in cross-layer knowledge distillation, which is verified in Section \ref{subsec:sem}.
	\item We present a theoretical connection between layer association weights and the classic Orthogonal Procrustes problem \cite{hurley1962procrustes,schonemann1966generalized} to further enlighten the underlying rationality of our approach.
	\item Extensive experiments on standard benchmarks with a large variety of settings based on popular network architectures demonstrate that SemCKD consistently generalizes better than state-of-the-art approaches.
\end{itemize}

\section{Related Work}
\label{sec:related}
\subsection{Knowledge Distillation}
Nowadays complex deep neural networks with millions of parameters or an ensemble of such models has achieved great success in various settings \cite{he2016deep,he2017mask,silver2017mastering,devlin2019bert,yang2019xlnet}, but it is still a huge challenge to efficiently deploy them in applications with limited computational and storage resources \cite{denil2013predicting,han2016deep}. There are some pioneering efforts, collectively known as \textit{model compression}, proposing a series of feasible solutions to this problem, such as knowledge distillation \cite{ba2014deep,hinton2015distilling}, low-rank approximation \cite{denil2013predicting,novikov2015tensor}, quantization \cite{gong2015compressing,polino2018model} and pruning \cite{han2015learning,li2017pruning}. Among these techniques, knowledge distillation is perhaps the easiest and the most hardware-friendly one for implementation due to its minimal modification to the regular training procedure. 

Given a lightweight student model, the aim of vanilla KD is to improve its generalization ability by training to match the predictions, or \textit{soft targets}, from a pre-trained cumbersome teacher model \cite{hinton2015distilling}. The intuitive explanation for the success of KD is that compared to discrete labels, fine-grained information among different categories in soft targets may provide extra supervision to improve the student model performance \cite{hinton2015distilling,rafael2019when}. A new interpretation for this improvement is that soft targets act as a learned label smoothing regularization to keep the student model from producing over-confident predictions \cite{yuan2020revisiting}. To save the expense of pre-training, some online variants have been proposed to explore cost-effective collaborative learning \cite{anil2018large,zhang2018deep,chen2020online}. By training a group of student models and encouraging each one to distill the knowledge in group-derived soft targets as well as ground-truth labels, these approaches could even surpass the vanilla KD in some teacher-student combinations \cite{zhang2018deep,chen2020online}.

\subsection{Feature-Map Distillation}
\label{subsec:related_fmd}
Rather than only formalizing knowledge in a highly abstract form like predictions, recent methods attempted to leverage information contained in intermediate layers by designing elaborate knowledge representations. 
A bunch of techniques have been developed for this purpose, such as aligning hidden layer responses called \textit{hints} \cite{romero2015fitnet}, encouraging the similar patterns to be elicited in the spatial attention maps \cite{zagoruyko2017paying} and maximizing the mutual information through variational inference \cite{ahn2019variational}. 
Since the spatial dimensions of feature maps or their transformations from teacher and student models usually vary, a straightforward step before the subsequent loss calculation is to match them by adding convolutional layers or pooling layers \cite{romero2015fitnet,zagoruyko2017paying,ahn2019variational}.
Some other parameter-free alternatives capture the transferred knowledge by crude pairwise similarity matrices \cite{tung2019similarity} or hybrid kernel formulations built on them to avoid mismatch, such as cosine-based and t-student transformations \cite{passalis2020heterogeneous}.  
With these pre-defined representations, all above approaches perform knowledge transfer with certain hand-crafted layer associations \cite{romero2015fitnet,tung2019similarity,zagoruyko2017paying,ahn2019variational,passalis2020heterogeneous,yue2020mgd}. Unfortunately, as pointed in the previous work \cite{passalis2020heterogeneous}, these hard associations would make the student model suffer from negative regularization, limiting the effectiveness of feature-map distillation. We will discuss this negative regularization in more depth in Section \ref{subsubsec:negative}.

Although a large number of approaches explored various knowledge representations over the past few years \cite{romero2015fitnet,tung2019similarity,zagoruyko2017paying,ahn2019variational,passalis2020heterogeneous,yue2020mgd}, very few have considered to leverage these representations in a better fashion. 
The most related approach to ours, which is proposed recently, works on channel or group-wise association within given teacher-student layer pairs \cite{yue2020mgd,guan2020differentiable}. The main difference is that we exploit feature maps from a more general cross-layer level rather than channel or group level.

Learning layer association has also been studied in transfer learning. A previous work learns the association weights between source and target networks by additional meta-networks with bi-level optimization \cite{jang2019learn}. Our approach differs from this one since feature maps of teacher-student layer pairs are both incorporated to derive the association weights while only those of the source network are used in \cite{jang2019learn}. Another advantage is that our association weights are trained end-to-end with the original student network, i.e. no need for carefully tuned bi-level optimization.

\subsection{Feature-Embedding Distillation}
\label{subsec:related_fed}
Feature embeddings are good substitutes for feature maps to distill knowledge in intermediate layers. 
Compared to high dimensional feature maps, these vectorized embeddings obtained at the penultimate layer are generally more tractable. Meanwhile, feature embeddings preserve more structural information than the final predictions used in the vanilla KD. 
Therefore, a variety of feature-embedding distillation approaches have been proposed recently \cite{passalis2018learning,park2019relational,liu2019knowledge,peng2019correlation,tian2020contrastive,xu2020knowledge}. In these approaches, a commonly used procedure is to organize all instances by a relational graph for knowledge transfer \cite{passalis2018learning,peng2019correlation,park2019relational,liu2019knowledge}. The main difference among them lies on how the edge weights are calculated. Typical choices include cosine kernel \cite{passalis2018learning}, truncated Gaussian RBF kernel \cite{peng2019correlation}, or combination of distance-wise and angle-wise potential functions \cite{park2019relational}. In contrast to pairwise transfer, some approaches formulate knowledge distillation from the perspective of contrastive learning to capture higher-order dependencies in the representation space \cite{tian2020contrastive,xu2020knowledge}.

Although our approach mainly focuses on feature-map distillation, it is also compatible with the state-of-the-art feature embedding distillation approach for further performance improvement.

\begin{figure*}
	\centering
	\includegraphics[width=0.9\textwidth]{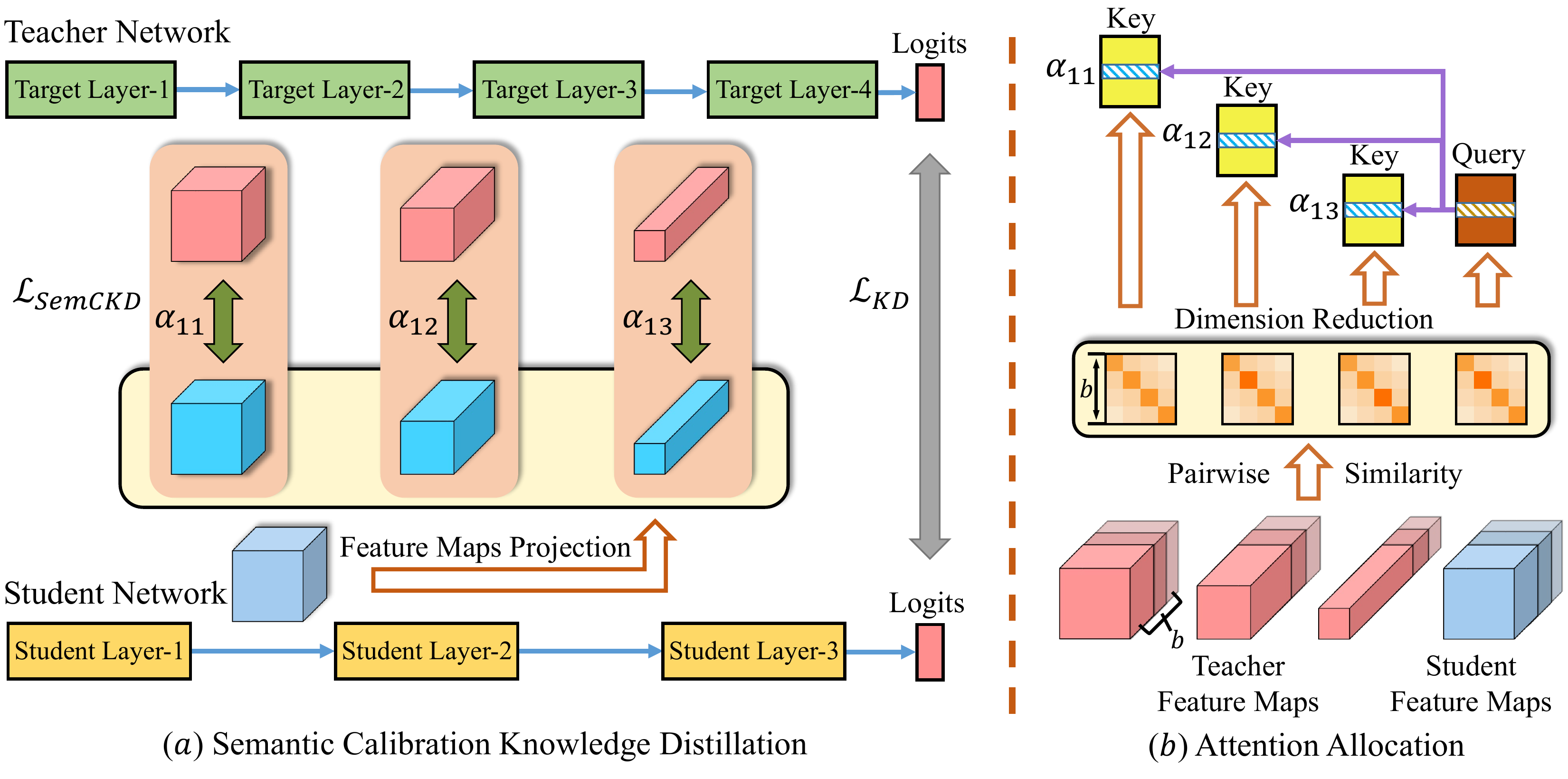}
	\caption{An overview of the proposed Semantic Calibration for Knowledge Distillation (SemCKD). ($\bm{a}$) Feature maps for a certain instance from the student layer-1 are projected into three individual forms to align with the spatial dimensions of those from associated target layers. The learned attention allocation adaptively helps the student model focus on the most semantic-related information for effective distillation. ($\bm{b}$) Pairwise similarities are first calculated between every stacked feature maps and then the attention weights are obtained by the proximities among generated \textit{query} and \textit{key} vectors.}
	\label{fig:model}
\end{figure*}

\section{Semantic Calibration for Distillation}

\subsection{Notations and Vanilla Knowledge Distillation}
In this section, we briefly recap the basic concepts of vanilla knowledge distillation as well as provide necessary notations for the following illustration.

Given a training dataset $\mathcal{D}=\{\left(\bm{x}_i,\bm{y}_i\right)\}_{i=1}^{N}$ consisting of $N$ instances from $K$ categories and a powerful teacher model pre-trained on the dataset $\mathcal{D}$, the goal of vanilla KD is reusing the same dataset to train another simple student model with cheaper computational and storage demand. 

For a mini-batch of instances with size $b$, we denote the outputs of each target layer $t_{l}$ and student layer $s_{l}$ as $F^{t}_{t_l}\in \mathbb{R}^{b\times c_{t_l}\times h_{t_l}\times w_{t_l}}$ and $F^{s}_{s_l} \in \mathbb{R}^{b\times c_{s_l}\times h_{s_l}\times w_{s_l}}$, respectively, where $c$ is the channel dimension, $h$ and $w$ are the spatial dimensions, superscript $t$ and $s$ reflect the corresponding teacher and student models. These outputs are also named as feature maps. $t_l/s_l$ ranges from $1$ to the total number of intermediate layers $t_L/s_L$. Note that $t_L$ and $s_L$ may be different especially when the teacher and student models adopt different architectures. The representations at the penultimate layer from teacher and student models are denoted as $F^{t}_{t_L+1}$ and  $F^{s}_{s_L+1}$, which are mainly used in feature-embedding distillation.  
Take the student model as an example, outputs of the last fully connected layer $g(\cdot)$ are known as logits
$\bm{g}^{s}_{i}=g(F^{s}_{s_L+1}[i]) \in \mathbb{R}^{K}$ and the predicted probabilities
are calculated with a softmax layer built on the logits, i.e., $\sigma(\bm{g}^{s}_{i}/T)$ with the temperature $T$ usually equals to $1$. The notation $F^{s}_{s_l}[i]$ denotes the output of student layer $s_{l}$ for the $i$-th instance and is a shorthand for $F^{s}_{s_l}[i,:,:,:]$. 

For classification tasks, in addition to regular cross entropy loss ($\mathcal{L}_{CE}$) between the predicted probabilities $\sigma(\bm{g}^{s}_{i})$ and the one-hot label $\bm{y}_i$ of each training sample,
classic KD \cite{hinton2015distilling} incorporates another alignment loss to encourage the minimization of Kullback-Leibler (KL) divergence \cite{kullback1951information}
between $\sigma(\bm{g}^{s}_{i}/T)$ and soft targets $\sigma(\bm{g}^{t}_{i}/T)$ predicted by the teacher model
\begin{equation}
	\label{eq:kd}
	\begin{aligned}
		\mathcal{L}_{KD_{i}}= \mathcal{L}_{CE}(\bm{y}_i, \sigma(\bm{g}^{s}_{i}))+T^2 \mathcal{L}_{KL}(\sigma(\bm{g}^{t}_{i}/T), \sigma(\bm{g}^{s}_{i}/T)),
	\end{aligned}
\end{equation}
where $T$ is a hyper-parameter and a higher $T$ leads to more considerable softening effect. The coefficient $T^2$ in the loss function is used to ensure that the gradient magnitude of the $\mathcal{L}_{KL}$ part keeps roughly unchanged when the temperature is larger than one \cite{hinton2015distilling}. 

\subsection{Feature-Map Distillation}
\label{subsec:fmd}
As mentioned earlier, feature maps of a teacher model are valuable for helping a student model achieve better performance. Recently proposed feature-map distillation approaches can be summarized as adding the following loss to Equation (\ref{eq:kd}) for each mini-batch with size $b$
\begin{equation}
	\label{eq:fmd}
	\begin{aligned}
		\mathcal{L}_{FMD}=\sum_{\left(s_{l},t_{l}\right)\in \mathcal{C}} \mathrm{Dist}&\left(\mathrm{Trans^{t}}\left(F^{t}_{t_l}\right), \mathrm{Trans^{s}}\left(F^{s}_{s_l}\right)\right),
	\end{aligned}
\end{equation}
leading to the overall loss as
\begin{equation}
	\label{eq:total}
	\mathcal{L}_{total} = \sum_{i=1}^{b} \mathcal{L}_{KD_{i}} + \beta \mathcal{L}_{FMD}.
\end{equation}
The functions $\mathrm{Trans^{t}(\cdot)}$ and $\mathrm{Trans^{s}(\cdot)}$ in Equation (\ref{eq:fmd}) are used to transform feature maps into particular representations.
The layer association set $\mathcal{C}$ of existing approaches are designed as one-pair selection \cite{romero2015fitnet,tung2019similarity} or one-to-one match \cite{zagoruyko2017paying,ahn2019variational,passalis2020heterogeneous,yue2020mgd}. 
As for one-pair selection, the candidate set $\mathcal{C}=\{(s_r,t_r)\}$ includes only one pair of a manually specified student layer $s_r$ and a teacher layer $t_r$, while for one-to-one match, the candidate set $\mathcal{C}=\{(1,1),...,\left(\mathrm{min}(s_{L},t_{L}),\mathrm{min}(s_{L},t_{L})\right)\}$ includes $|\mathcal{C}|=\mathrm{min}(s_{L},t_{L})$ elements.
With these associated layer pairs, the loss in Equation (\ref{eq:fmd}) is calculated by a certain distance function $\mathrm{Dist(\cdot, \cdot)}$. A common practice is to use Mean-Square-Error (MSE) \cite{romero2015fitnet} sometimes with a normalization pre-processing \cite{zagoruyko2017paying,tung2019similarity,yue2020mgd}.
The hyper-parameter $\beta$ in Equation (\ref{eq:total}) is used to balance the two individual loss terms. 
The detailed comparison of several representative approaches is provided in Table \ref{Table:comp}. 

As shown in Table \ref{Table:comp}, FitNet \cite{romero2015fitnet} adds a convolutional transformation, i.e., $\mathrm{Conv(\cdot)}$ on top of a certain intermediate layer of the student model,  while keeping feature maps from the teacher model unchanged by identity transformation $\mathrm{I(\cdot)}$. 
AT \cite{zagoruyko2017paying} encourages the student to mimic spatial attention maps of the teacher by channel attention. VID \cite{ahn2019variational} formulates knowledge transfer as Mutual Information (MI) maximization between feature maps and expresses the distance function as Negative Log-Likelihood (NLL). 
The transferred knowledge can also be captured by crude similarity matrices or hybrid kernel transformations build on them \cite{tung2019similarity,passalis2020heterogeneous}.
MGD \cite{yue2020mgd} makes the channel dimensions of feature maps in candidate layer pairs become matched by an assignment module $\rho(\cdot)$. Note that almost twice as much computational and storage resources are needed for HKD \cite{passalis2020heterogeneous}. Since it first transfers feature embeddings 
from the teacher to an auxiliary model, which follows the same architecture as the student but with twice parameters per layer, then performs feature-map distillation from the auxiliary to the student model. 

\begin{table}
	\caption{Comparison of different approaches.}
	\label{Table:comp}
	\centering
	\renewcommand\arraystretch{1.2} 
	\resizebox{0.98\columnwidth}{!}{
		\begin{tabular}{cccc}
			\toprule
			Methods & $\mathrm{Trans^t(\cdot)}$ & $\mathrm{Trans^s(\cdot)}$ & $\mathrm{Dist(\cdot)}$  \\
			\midrule
			FitNet \cite{romero2015fitnet} 
			& $\mathrm{I(\cdot)}$ & $\mathrm{Conv(\cdot)}$ & $\mathrm{MSE}$ \\
			AT \cite{zagoruyko2017paying}    
			& Channel Attention & Channel Attention & $\mathrm{MSE}$ \\
			SP \cite{tung2019similarity}     
			& Similarity Matrices & Similarity Matrices & $\mathrm{MSE}$ \\
			VID \cite{ahn2019variational}
			& Variational MI & Variational MI & $\mathrm{NLL}$  \\
			HKD \cite{passalis2020heterogeneous}  
			& $\mathrm{Kernel(\cdot)}$ & $\mathrm{Kernel(\cdot)}$ & $\mathrm{KL}$  \\
			MGD \cite{yue2020mgd} 
			& $\rho(\cdot)$ & $\mathrm{I(\cdot)}$ & $\mathrm{MSE}$  \\
			\bottomrule
		\end{tabular}
	}
\end{table}

All of the above approaches perform knowledge transfer based on fixed associations between assigned teacher-student layer pairs, which may cause the loss of useful information. Take one-to-one match as an example, extra layers are discarded when $s_{L}$ and $t_{L}$ differ. Moreover, forcing feature maps from the same layer depths to be aligned may result in suboptimal associations, since a better choice for the student layer could from different or multiple target teacher layers. 

To solve these problems, we propose a new cross-layer knowledge distillation to promote the exploitation of feature-map representations. Based on our learned layer association weights, simple convolutional $\mathrm{Trans^t(\cdot)}$, $\mathrm{Trans^s(\cdot)}$ and $\mathrm{MSE}$ distance are enough to achieve state-of-the-art results.
\subsection{Semantic Calibration Formulation}
In our approach, each student layer is automatically associated with those semantic-related target layers by attention allocation, as illustrated in Figure \ref{fig:model}. 
Training with soft association weights encourages the student model to collect and integrate multi-layer information to obtain a more suitable regularization. Moreover, SemCKD is readily applicable to situations where the numbers of candidate layers in teacher and student models are different. 

The learned association set $\mathcal{C}$ in SemCKD is denoted as 
\begin{equation}
	\label{eq:can}
	\mathcal{C}=\{(s_{l},t_{l}) \,|\,\, \forall\,s_{l}\in [1,...,s_{L}],\, t_{l} \in[1,...,t_{L}]\},
\end{equation}
with the corresponding weight satisfies $\sum_{t_{l}=1}^{t_{L}}\bm{\alpha}_{(s_{l},t_{l})}=\bm{1},$ $\forall\,s_{l}\in [1,...,s_{L}]$. The weights for a mini-batch of instances $\bm{\alpha}_{(s_{l},t_{l})}\in \mathbb{R}^{b\times 1}$ represent the extent to which the target layer $t_{l}$ is attended in deriving the semantic-aware guidance for the student layer $s_{l}$. Each instance will hold its own association weight $\alpha_{(s_{l},t_{l})}[i]$ for the layer pair $(s_{l},t_{l})$, which is calculated by function $\mathcal{F}(\cdot, \cdot)$ of given feature maps
\begin{equation}
	\label{eq:alpha}
	\alpha_{(s_{l},t_{l})}[i]=\mathcal{F}(F^{s}_{s_l}[i], F^{t}_{t_l}[i]).
\end{equation}
The detailed design of $\mathcal{F}(\cdot, \cdot)$ will be elaborated later.
Given these association weights, feature maps of each student layer are projected into $t_{L}$ individual tensors to align with the spatial dimensions of those from each target layer:
\begin{equation}
	\label{eq:proj}
	F^{s^{\prime}}_{t_l}=\mathrm{Proj}\left(F^{s}_{s_l} \in \mathbb{R}^ {b\times c_{s_l}\times h_{s_l}\times w_{s_l}}, t_{l}\right), t_{l} \in [1,...,t_{L}],
\end{equation}
with $F^{s^{\prime}}_{t_l}\in \mathbb{R}^{b\times c_{t_l}\times h_{t_l}\times w_{t_l}}$. Each function $\mathrm{Proj(\cdot, \cdot)}$ includes a stack of three layers with $1\times1$, $3\times3$ and $1\times1$ convolutions to meet the demand of capability for effective transformation\footnote{In practice, a pooling operation is first used to reconcile the height and weight dimensions of $F^{t}_{s_l}$ and $F^{s}_{s_l}$ before projections to reduce computational consumption.}.

\textbf{Loss function.} For a mini-batch of instances with size $b$, the trainable student model and fixed teacher model will produce several feature maps across multiple layers, i.e., $F^{s}_{s_1}, ..., F^{s}_{s_L}$ and $F^{t}_{t_1}, ..., F^{t}_{t_L}$, respectively. After attention allocation and dimensional projections, the feature-map distillation loss ($\mathcal{L}_{FMD}$) of SemCKD is obtained by simply using $\mathrm{MSE}$ for distance function
\begin{equation}
	\label{eq:semCKD}
	\small
	\begin{aligned}
		\mathcal{L}_{SemCKD}
		&=\sum_{\left(s_{l},t_{l}\right)\in \mathcal{C}} \mathrm{Dist}\left(\mathrm{Trans^t}(F^{t}_{t_l}), \mathrm{Trans^s}(F^{s}_{s_l}, t_{l}), \bm{\alpha}_{(s_{l},t_{l})} \right)\\
		&=\sum_{s_{l}=1}^{s_{L}}\sum_{t_{l}=1}^{t_{L}}\sum_{i=1}^{b}\alpha_{(s_{l},t_{l})}[i]\, \mathrm{Dist}\left(F^{t}_{t_l}[i], \mathrm{Proj}\left(F^{s}_{s_l}[i], t_{l}\right) \right)\\
		&=\sum_{s_{l}=1}^{s_{L}}\sum_{t_{l}=1}^{t_{L}}
		\sum_{i=1}^{b}\alpha_{(s_{l},t_{l})}[i]\, \mathrm{MSE}\left(F^{t}_{t_l}[i], F^{s^{\prime}}_{t_{l}}[i]\right),
	\end{aligned}
\end{equation}
where feature maps from each student layer are transformed by a projection function as Equation (\ref{eq:proj}) while those from the target layers remain unchanged by identity transformation $\mathrm{Trans^{t}(\cdot)}=\mathrm{I(\cdot)}$. Equipped with the learned layer association weights, the total loss is aggregated by a weighted summation of each individual distance among the feature maps from candidate teacher-student layer pairs. Note that FitNet \cite{romero2015fitnet} is a special case of SemCKD by fixing $\alpha_{(s_{l},t_{l})}[i]$ to $1$ for certain $(s_{l},t_{l})$ layer pair and $0$ for the rest.

\textbf{Attention Allocation.}
Now we provide concrete steps to obtain the association weights. As pointed in previous works, feature representations contained in a trained neural network are progressively more abstract as the layer depth increases \cite{bengio2013representation,zeiler2014visualizing,goodfellow2016deep}.
Thus, layer semantics in teacher and student models 
usually varies. Existing hand-crafted strategies, which did not take this factor into consideration, may not suffice due to negative effects caused by semantic mismatched layers \cite{passalis2020heterogeneous}. 
To further improve the performance of feature-map distillation, each student layer had better associate with the most semantic-related target layers to derive its own regularization. 

Layer association based on attention mechanism provides a potentially feasible solution for this goal. 
Based on the observation that feature maps produced by similar instances probably become clustered at separate granularity in different intermediate layers \cite{tung2019similarity}, we regard the proximity of pairwise similarity matrices as a good measurement of the inherent semantic similarity. These matrices are calculated as
\begin{equation}
	\label{eq:adj}
	A^{s}_{s_{l}} = \mathrm{R}(F^{s}_{s_{l}})\cdot \mathrm{R}(F^{s}_{s_{l}})^T \quad A^{t}_{t_{l}} = \mathrm{R}(F^{t}_{t_{l}})\cdot \mathrm{R}(F^{t}_{t_{l}})^T,
\end{equation}
where $\mathrm{R(\cdot)}: \mathbb{R}^{b\times c\times h\times w}\mapsto \mathbb{R}^{b\times (c\cdot h\cdot w)}$ is a reshaping operation, and therefore $ A^{s}_{s_{l}}$ and $A^{t}_{t_{l}}$ are $b\times b$ matrices.

Inspired by the self-attention framework \cite{vaswani2017attention}, we separately project the pairwise similarity matrices of each student layer and associated target layers into two subspaces by a Multi-Layer Perceptron ($\mathrm{MLP}$) to alleviate the effect of noise and sparseness. For the $i$-th instance 
\begin{equation}
	\label{eq:mlp}
	Q_{s_{l}}[i]=\mathrm{MLP}_{Q}(A^{s}_{s_{l}}[i]) \quad
	K_{t_{l}}[i]=\mathrm{MLP}_{K}(A^{t}_{t_{l}}[i]).
\end{equation}
The parameters of $\mathrm{MLP}_{Q}(\cdot)$ and $\mathrm{MLP}_{K}(\cdot)$ are learned during training to generate \textit{query} and \textit{key} vectors and shared by all instances.
Then, $\alpha_{(s_{l},t_{l})}[i]$ is calculated as follows
\begin{equation}
	\label{eq:att}
	\alpha_{(s_{l},t_{l})}[i]=\mathcal{F}(A^{s}_{s_l}[i], A^{t}_{t_l}[i])=
	\frac{e^{Q_{s_{l}}[i]\cdot K_{t_{l}}[i]^{\mathrm{T}}}}
	{\sum_{j} e^{Q_{s_{l}}[i]\cdot K_{t_{j}}[i]^{\mathrm{T}}}}.
\end{equation}
where $\mathcal{F}(\cdot, \cdot)$ for calculating the attention weights is named as \textit{Embeded Gaussian} operation in the non-local block \cite{wang2018non-local}. Attention-based allocation provides a possible way to suppress negative effects caused by layer mismatch and integrate positive guidance from multiple target layers, which is supported by the theoretical discussion in Section \ref{subsec:theory} and empirical evidence in Section \ref{subsec:sem}. Although our proposed SemCKD distills only the knowledge contained in intermediate layers, its performance can be further boosted by incorporating additional regularization, e.g., feature-embedding distillation.
The full training procedure with the proposed semantic calibration formulation is summarized in Algorithm \ref{alg:Framwork}.
\subsection{Theoretical Insights}
\label{subsec:theory}
In this section, we present a theoretical connection between our proposed attention allocation and the famous Orthogonal Procrustes problem \cite{hurley1962procrustes,schonemann1966generalized,golub2013matrix}.

The key design of attention allocation is function $\mathcal{F}(\cdot, \cdot)$, which measures the strength of  association weights and should satisfy non-negative constrain. In our approach, this constrain is naturally ensured since each used feature maps have passed a ReLU activation function ($\max (0, x)$).
Thus, we can safely replace the \textit{Embedded Gaussian} with a computation-efficient \textit{Dot Production} operation in Equation (\ref{eq:att}), which has achieved empirical success in the previous work \cite{wang2018non-local}. Additional, we omit the constant normalization term suggested by \cite{wang2018non-local} in the following discussion for simplicity.

\begin{algorithm}[htbp] 
	\caption{Semantic Calibration for Distillation.}
	\label{alg:Framwork} 
	\begin{algorithmic}[1] 
		\REQUIRE
		Training dataset $\mathcal{D}=\{(\bm{x}_i, \bm{y}_i)\}_{i=1}^{N}$; A pre-trained teacher model with parameter $\theta^t$; A student model with randomly initialized parameters $\theta^s$; 
		\ENSURE A well-trained student model; 
		\WHILE{$\theta^s$ is not converged}
		\STATE 
		Sample a mini-batch $\mathcal{B}$ with size $b$ from $\mathcal{D}$.
		\STATE
		Forward propagation $\mathcal{B}$ into $\theta^t$ and $\theta^s$ to obtain intermediate presentations $F^{t}_{t_{l}}$ and $F^{s}_{s_{l}}$ across layers.
		\STATE
		Construct pairwise similarity matrices $A^{t}_{t_{l}}$ and $A^{s}_{s_{l}}$ as Equation (\ref{eq:adj}).
		\STATE
		Perform attention allocation as Equation (\ref{eq:mlp}-\ref{eq:att}).
		\STATE
		Align feature maps by projections as Equation (\ref{eq:proj}).
		\STATE
		Update parameters $\theta^s$ by backward propagation the gradients of the loss in Equation (\ref{eq:total}) and Equation (\ref{eq:semCKD}).
		\ENDWHILE
	\end{algorithmic} 
\end{algorithm}

Suppose that the $\mathrm{MLP}(\cdot)$ in Equation (\ref{eq:mlp}) equals to identity transformation\footnote{A similar result will be obtained for the case with the original $\mathrm{MLP}(\cdot)$.}, and the attention weight $w_{(s_l,t_l)}$ for candidate layer pair $(s_{l},t_{l})$ is calculated by averaging the weights of all instances
\begin{equation}
	\label{eq:weight}
	\begin{aligned}
		w_{(s_l,t_l)}&=\frac{1}{b}\sum_{i=1}^{b}\alpha_{(s_{l},t_{l})}[i]=\frac{1}{b}\sum_{i=1}^{b}Q_{s_l}[i]\cdot K_{t_l}[i]^{\mathrm{T}} \\
		&=\frac{1}{b}\mathrm{Tr}(Q_{s_l}\cdot K_{t_l}^{\mathrm{T}})=\frac{1}{b}
		\mathrm{Tr}(A^{s}_{s_l}\cdot {A_{t_l}^{t}}^{\mathrm{T}})	\\
		&=\frac{1}{b}\|\mathrm{R}(F^{t}_{t_l})^{\mathrm{T}}\cdot \mathrm{R}(F^{s}_{s_l})\|_{\mathrm{F}}^2, 
	\end{aligned}
\end{equation}
where $\mathrm{R}(F^{t}_{t_l}), \mathrm{R}(F^{s}_{s_l})\in \mathbb{R}^{b \times (c\cdot h\cdot w)}$\footnote{We suppose the vectorized feature maps from teacher and student models sharing the same dimension $(c\cdot h\cdot w)$ in what follows.}. From the last Equation of (\ref{eq:weight}), we can see that $w_{(s_l,t_l)}$ actually measures the extent to which the extracted feature maps from teacher and student models are shared or similar. We then establish a connection between this weight and the optimal value of the Orthogonal Procrustes problem.
\begin{lemma}
	\label{lemma:norm}
	Given a matrix $X\in \mathbb{R}^{m\times n}$ with the $i$-th largest singular value denoted as $\sigma_i$, and $i\in[1, p]$, $ p=\mathrm{min}(m, n)$. The Frobenius norm $\|\mathrm{X}\|_{\mathrm{F}}=\sqrt{\sum_{i=1}^{p}\sigma_i^2}$ and nuclear norm $\|\mathrm{X}\|_{*}=\sum_{i=1}^{p}\sigma_i$ will bound each other
\end{lemma}
\begin{equation}
	\label{eq:ineq}
	\begin{aligned}
		\frac{1}{\sqrt{p}}\|\mathrm{X}\|_{*} \leq\|\mathrm{X}\|_{\mathrm{F}} \leq\|\mathrm{X}\|_{*} \leq \sqrt{p}\|\mathrm{X}\|_{\mathrm{F}}.
	\end{aligned}
\end{equation}
\textit{Proof.}
Since $\sigma_i \ge 0$, the inequality 
$\left(\sum_{i=1}^{p}\sigma_i\right)^2 = \sum_{i=1}^{p} \sigma_i^2 + 2 \sum_{i=2}^{p}\sum_{j=1}^{i-1}\sigma_i \sigma_j \ge \sum_{i=1}^{p} \sigma_i^2$
always holds, which implies that $\|\mathrm{X}\|_{\mathrm{F}} \leq\|\mathrm{X}\|_{*}$. Based on the Cauchy-Schwarz inequality 
$\left(\sum_{i=1}^{p}\sigma_i\delta_i\right)^2\leq\left(\sum_{i=1}^{p}\sigma_i^2\right)\left(\sum_{i=1}^{p}\delta_i^2\right)$ and let $\delta_i \equiv 1$, we can obtain that $\|\mathrm{X}\|_{*} \leq \sqrt{p}\|\mathrm{X}\|_{\mathrm{F}}$.
\qed

\textbf{The Orthogonal Procrustes problem.} This optimization problem seeks an orthogonal transformation $O$ to make the matrix $A\in \mathbb{R}^{m\times n}$ align with the matrix $B\in \mathbb{R}^{m\times n}$ such that Frobenius norm of the residual error matrix $E$ is minimized \cite{hurley1962procrustes,schonemann1966generalized}. The mathematical expression is presented as follows
\begin{equation}
	\label{eq:op}
	\begin{aligned}
		E=\mathop{\min}_O ~||B - AO||_{\mathrm{F}}^2\quad~ \mbox{s.t.} ~~O^{\mathrm{T}}O = I_{n}.
	\end{aligned}
\end{equation}
Since $||B - A O||_{\mathrm{F}}^2 = ||B||_{\mathrm{F}}^2 + ||A||_{\mathrm{F}}^2 - 2\mathrm{Tr}(B^{\mathrm{T}}AO)$ and $||B||_{\mathrm{F}}^2$, $||A||_{\mathrm{F}}^2$ are constants, the objective of Equation (\ref{eq:op}) can be rewritten as an equivalent one
\begin{equation}
	\label{eq:op_new}
	E^\prime = \mathop{\max}_O \mathrm{Tr}(B^\mathrm{T}AO)\quad~\mbox{s.t.}~~O^\mathrm{T}O = I_{n} .
\end{equation}
By applying the method of Lagrange multipliers \cite{bertsekas2014constrained}, we can convert this constrained problem into an unconstrained one to find the solution $\hat O = U V^\text{T}$, where $U$ and $V$ are obtained by singular value decomposition $A^\mathrm{T}B= U\Sigma V^\mathrm{T}$ \cite{golub2013matrix}. Thus, the maximum objective of Equation (\ref{eq:op_new}) is
\begin{equation}
	\label{eq:nuclear}
	E^\prime =\mathrm{Tr}(B^\mathrm{T}A\hat{O}) = \mathrm{Tr}(\Sigma) = ||A^\mathrm{T}B||_{*} = ||B^\mathrm{T}A||_{*}.
\end{equation}

Our crucial observation is that if we interpret ${E^\prime}^2$ as a measurement of inherent semantic similarity for feature maps, our designed layer association weights will naturally achieve semantic calibration.

Given vectorized feature maps from the teacher and student models $\mathrm{R}(F^{t}_{t_l})$ and $\mathrm{R}(F^{s}_{s_l})$, the irreducible error of orthogonal transformation from $\mathrm{R}(F^{s}_{s_l})$ to $\mathrm{R}(F^{t}_{t_l})$, as discussed in the above, is related to ${E^\prime}^2$, or $\|\mathrm{R}(F^{t}_{t_l})^\mathrm{T}\cdot \mathrm{R}(F^{s}_{s_l})\|_{\mathrm{*}}^2$. 
The larger ${E^\prime}^2$ means that these two representations are more similar from the perspective of orthogonal transformation.
Since the lemma \ref{lemma:norm} implies that increasing $||X||_{\mathrm{*}}$ tends to increase $||X||_{\mathrm{F}}$ and vice versa, 
$\|R(F^{t}_{t_l})^\mathrm{T}\cdot \mathrm{R}(F^{s}_{s_l})\|_{\mathrm{F}}^2$ will tend to be larger if ${E^\prime}^2$ becomes larger.
That is to say, attention allocation achieve semantic calibration by assigning larger weight $w_{(s_l,t_l)}$ for a certain teacher-student layer pair once the generated feature maps of student layer $\mathrm{R}(F^{s}_{s_l})$ can be orthogonally transformed into those of the associated target layer $\mathrm{R}(F^{t}_{t_l})$ with relatively lower error.
\subsection{Softening Attention}


A simple yet effective variant to further improve the performance of SemCKD is presented in this section. 
Similar to softening predicted probabilities in the vanilla KD as Equation (\ref{eq:kd}), we find that softening attention in Equation (\ref{eq:att}) is also helpful. But the reason behind these two seemingly similar steps are completely different. The former is to capture more structural information among different classes while the latter is to smooth the gradient direction guided by different target layers for each student layer in training.

In this variant SemCKD$_{\tau}$, we replace Equation (\ref{eq:att}) as 
\begin{equation}
	\label{eq:att_s}
	\alpha_{(s_{l},t_{l})}[i]=\frac{e^{\left(Q_{s_{l}}[i]\cdot K_{t_{l}}[i]^{\mathrm{T}}\right)/\tau}}
	{\sum_{j} e^{\left(Q_{s_{l}}[i]\cdot K_{t_{j}}[i]^{\mathrm{T}}\right)/\tau}}.
\end{equation}

Note that $\tau$ is set as a constant to counteract extremely small gradient during training in the original self-attention framework \cite{vaswani2017attention}, 
but it is regarded as a new hyper-parameter in SemCKD$_{\tau}$. We will discuss the improvement over SemCKD with various $\tau$ in more detail in Section \ref{subsec:softening}.

\section{Experiments}
We conduct extensive experiments in this section to demonstrate the effectiveness of our proposed approach.
In addition to comparing with some representative feature-map distillation approaches, we provide results to support and explain the success of our semantic calibration strategy in helping student models obtain a proper regularization through several carefully designed experiments (Section \ref{subsec:sem}) and ablation studies (Section \ref{subsec:ab}).
Moreover, we provide a variant SemCKD$_{\tau}$ for further performance improvement and show that SemCKD generalizes well in different scenarios such as transfer learning, few-shot learning as well as noisy-label learning. SemCKD is also compatible with the state-of-the-art feature-embedding distillation technique to achieve better results. Finally, sensitivity analysis to the hyper-parameter $\beta$ is reported.

\begin{table*}[htbp]
	\centering
	\caption{Top-1 test accuracy of \textit{feature-map distillation} approaches on CIFAR-100.}
	\label{Table:CIFAR-100-1}
	\resizebox{0.9\textwidth}{!}{
		\begin{tabular}{c|cccccc}
			\toprule
			\multirow{2}{*}{Student} & VGG-8 & VGG-13 & ShuffleNetV2 & ShuffleNetV2 & MobileNetV2 & ShuffleNetV1 \\
			& 70.46 $\pm$ 0.29 & 74.82 $\pm$ 0.22 & 72.60 $\pm$ 0.12 & 72.60 $\pm$ 0.12 & 65.43 $\pm$ 0.29 & 71.36 $\pm$ 0.25  \\
			\midrule
			KD \cite{hinton2015distilling} & 72.73 $\pm$ 0.15 & 77.17 $\pm$ 0.11 & 75.60 $\pm$ 0.21 & 75.49 $\pm$ 0.24 & 68.70 $\pm$ 0.22 & 74.30 $\pm$ 0.16 \\
			FitNet \cite{romero2015fitnet} & 72.91 $\pm$ 0.18 & 77.06 $\pm$ 0.14 & 75.44 $\pm$ 0.11 & 75.82 $\pm$ 0.22 & 68.64 $\pm$ 0.12 & 74.52 $\pm$ 0.03 \\
			AT \cite{zagoruyko2017paying} & 71.90 $\pm$ 0.13 & 77.23 $\pm$ 0.19 & 75.41 $\pm$ 0.10 & 75.91 $\pm$ 0.14 & 68.79 $\pm$ 0.13 & 75.55 $\pm$ 0.19 \\
			SP \cite{tung2019similarity} & 73.12 $\pm$ 0.10 & 77.72 $\pm$ 0.33 & 75.54 $\pm$ 0.18 & 75.77 $\pm$ 0.08 & 68.48 $\pm$ 0.36 & 74.69 $\pm$ 0.32 \\
			VID \cite{ahn2019variational} & 73.19 $\pm$ 0.23 & 77.45 $\pm$ 0.13 & 75.22 $\pm$ 0.07 & 75.55 $\pm$ 0.18 & 68.37 $\pm$ 0.24 & 74.76 $\pm$ 0.22 \\
			HKD \cite{passalis2020heterogeneous} & 72.63 $\pm$ 0.12 & 76.76 $\pm$ 0.13 & 76.24 $\pm$ 0.09 & 76.64 $\pm$ 0.05 & 69.23 $\pm$ 0.16 & 75.89 $\pm$ 0.09 \\
			MGD \cite{yue2020mgd} & 72.39 $\pm$ 0.16 & 77.17 $\pm$ 0.22 & 75.74 $\pm$ 0.21 & 75.78 $\pm$ 0.29 & 68.55 $\pm$ 0.14 & --- \\
			\midrule
			SemCKD & \textbf{75.27 $\pm$ 0.13} & \textbf{79.43 $\pm$ 0.02} & \textbf{76.39 $\pm$ 0.12} & \textbf{77.62 $\pm$ 0.32} & \textbf{69.61 $\pm$ 0.05} & \textbf{76.31 $\pm$ 0.20} \\
			\midrule
			\multirow{2}{*}{Teacher} & ResNet-32x4 & ResNet-32x4 & VGG-13 & ResNet-32x4 & WRN-40-2 & ResNet-32x4  \\
			& 79.42 & 79.42 & 74.64 & 79.42 & 75.61 & 79.42 \\
			\bottomrule
		\end{tabular}
	}
\end{table*}     

\subsection{Experimental Setup}
\subsubsection{Dataset} 
In this paper, four popular datasets including CIFAR-100 \cite{krizhevsky2009learning}, STL-10 \cite{coates2011an}, Tiny-ImageNet \cite{torralba2008million} and ImageNet \cite{russakovsky2015imagenet} are used to conduct a series of tasks on image classification, transfer learning, few-shot learning and noisy-label learning. 
A standard preprocessing procedure is adopted as previous works \cite{he2016deep,tian2020contrastive}, that is, all images are normalized by channel means and standard deviations. 

The CIFAR-100 dataset
contains 50,000 training samples and 10,000 testing samples, which are drawn from 100 fine-grained classes. Each training image is padded by 4 pixels on each size and extracted as a randomly sampled 32x32 crop or its horizontal filp for data augmentation.
The ImageNet dataset coming from ILSVRC-2012
is a more challenging for large-scale classification. It consists of about 1.3 million training images and 50,000 validation images from 1,000 classes. Each training image is extracted as a randomly sampled 224x224 crop or its horizontal filp without any padding operation.
The following two datasets are used for transfer learning task where all images are downsampled to 32x32. The STL-10 dataset
contains 5,000 training images and 8,000 testing images from 10 classes.
The Tiny-ImageNet
contains 100,000 training images and 10,000 validation images from 200 classes.
\subsubsection{Network Architectures}
A large variety of teacher-student combinations based on popular networks are used for evaluation \cite{simonyan2014very,he2016deep,zagoruyko2016wide,sandler2018mobile,zhang2018shuffle,ma2018shuffle}. 
The number behind ``VGG-'' or ``ResNet-'' represents the depth of networks. ``WRN-d-w'' represents wide-ResNet with depth $d$ and width factor $w$. The number behind ``x'' in ``ResNet-8x4'', ``ResNet-32x4'', ``ResNet-34x4'' or ``ShuffleNetV2x0.5'' indicates that the number of filters in each layer is expanded or contracted with the certain factor.

\subsubsection{Compared Approaches}
Three kinds of knowledge distillation approaches based on transferring knowledge from different positions are compared through our paper, which are listed as follows

\begin{itemize}
	\item \textit{Logits distillation} includes the vanilla KD \cite{hinton2015distilling}.
\end{itemize}

\begin{itemize}
	\item \textit{Feature-map distillation} includes FitNet \cite{romero2015fitnet}, AT \cite{zagoruyko2017paying}, SP \cite{tung2019similarity}, VID \cite{ahn2019variational}, HKD \cite{passalis2020heterogeneous} and MGD \cite{yue2020mgd}. The detailed discussion is presented in Section \ref{sec:related} and \ref{subsec:fmd}. 
\end{itemize}

\begin{itemize}
	\item \textit{Feature-embedding distillation} includes PKT \cite{passalis2018learning}, RKD \cite{park2019relational}, IRG \cite{liu2019knowledge}, CC \cite{peng2019correlation} and CRD \cite{tian2020contrastive}. The detailed discussion is presented in Section \ref{sec:related}.
\end{itemize}

\begin{table*}[htbp]
	\centering
	\caption{Top-1 test accuracy of \textit{feature-map distillation} approaches on CIFAR-100.}
	\label{Table:CIFAR-100-2}
	\resizebox{0.9\textwidth}{!}{
		\begin{tabular}{c|cccccc}
			\toprule
			\multirow{2}{*}{Student} & WRN-16-2 & ShuffleNetV1 & MobileNetV2 &  MobileNetV2 & VGG-8 & ResNet-8x4 \\
			& 73.41 $\pm$ 0.27 & 71.36 $\pm$ 0.25 & 65.43 $\pm$ 0.29  & 65.43 $\pm$ 0.29 & 70.46 $\pm$ 0.29 & 73.09 $\pm$ 0.30 \\
			\midrule
			KD \cite{hinton2015distilling} & 74.87 $\pm$ 0.09 & 75.13 $\pm$ 0.19 & 68.39 $\pm$ 0.32 & 69.70 $\pm$ 0.15 & 73.38 $\pm$ 0.05 &  74.42 $\pm$ 0.05 \\
			FitNet \cite{romero2015fitnet} & 74.87 $\pm$ 0.14 & 75.28 $\pm$ 0.21 & 68.28 $\pm$ 0.21 & 69.33 $\pm$ 0.23 & 73.63 $\pm$ 0.11 & 74.32 $\pm$ 0.08 \\
			AT \cite{zagoruyko2017paying} & 75.34 $\pm$ 0.08 & 76.39 $\pm$ 0.05 & 67.15 $\pm$ 0.16 & 69.37 $\pm$ 0.26 & 73.51 $\pm$ 0.08 &  75.07 $\pm$ 0.03 \\
			SP \cite{tung2019similarity} & 74.79 $\pm$ 0.05 & 75.92 $\pm$ 0.29 & 67.20 $\pm$ 0.71 & 70.04 $\pm$ 0.27 & 73.53 $\pm$ 0.23 & 74.29 $\pm$ 0.07 \\
			VID \cite{ahn2019variational} & 75.27 $\pm$ 0.08 & 75.97 $\pm$ 0.25 & 67.94 $\pm$ 0.24 & 69.92 $\pm$ 0.30 & 73.63 $\pm$ 0.07 &  74.55 $\pm$ 0.10 \\
			HKD \cite{passalis2020heterogeneous} & 74.99 $\pm$ 0.08 & 75.98 $\pm$ 0.09 & 68.88 $\pm$ 0.08 & 69.97 $\pm$ 0.21 & 73.06 $\pm$ 0.24 & 74.86 $\pm$ 0.21 \\
			MGD \cite{yue2020mgd} & 74.79 $\pm$ 0.18 & --- & 67.34 $\pm$ 0.41 & 69.68 $\pm$ 0.24 & 73.47 $\pm$ 0.24 & 74.36 $\pm$ 0.16 \\
			\midrule
			SemCKD & \textbf{76.24 $\pm$ 0.21} & \textbf{76.83 $\pm$ 0.27}&\textbf{69.00 $\pm$ 0.19} & \textbf{70.24 $\pm$ 0.12} &  \textbf{74.43 $\pm$ 0.25} & \textbf{76.23 $\pm$ 0.04} \\
			\midrule
			\multirow{2}{*}{Teacher} & ResNet-32x4 & WRN-40-2 & ResNet-32x4 & ShuffleNetV2 & VGG-13 & ResNet-32x4 \\
			& 79.42 & 75.61 & 79.42 & 72.60 & 74.64 & 79.42 \\
			\bottomrule
		\end{tabular}
	}
\end{table*}     
\subsubsection{Training Details} 

We use stochastic gradient descent with Nesterov momentum and set the momentum to 0.9 in all experiments. 
For CIFAR-100, we follow the setting of CRD \cite{tian2020contrastive}, that is, the initial learning rate is set as 0.01 for MobileNetV2, ShuffleNetV1/V2 and 0.05 for other architectures. 
All models are trained for 240 epochs and the learning rate is divided by 10 at 150, 180 and 210 epochs. We set the mini-batch size to 64 and the weight decay to $5\times10^{-4}$. All results are reported in means (standard deviations) over 4 trials. For ImageNet, the initial learning rate is 0.1 and divided by 10 at 30 and 60 of the total 90 training epochs. We set the mini-batch size to 256 and the weight decay to $1\times 10^{-4}$. All results are reported in a single trial.
We set the hyper-parameter $\beta$ of SemCKD to 400 and the temperature $T$ for the vanilla KD loss in Equation (\ref{eq:kd}) to $4$ throughout this paper. The detailed implementation could be found in our open-source code. We regard the building blocks of teacher and student networks as \textit{target layer} and \textit{student layer} in practice, which are significantly less than the total number of neural networks.

\subsubsection{Evaluation Metrics}
Besides the widely used \textit{Top-1 test accuracy (\%)} for performance evaluation, we adopt another metric named as \textit{Relative Improvement} ($RI$) and its variant \textit{Average Relative Improvement} ($ARI$) to obtain an intuitive sense about the quantitative improvement \cite{tian2020contrastive,chen2021cross}.
\begin{equation}
	\small
	ARI = \frac{1}{M}\sum_{i=1}^{M} \underbrace{\frac{Acc^{i}_{SemCKD}-Acc^{i}_{FMD}}{Acc^{i}_{FMD}-Acc^{i}_{STU}} \times 100\%}_{RI},
\end{equation} 
where $M$ is the number of combinations and $Acc^{i}_{SemCKD}$, $Acc^{i}_{FMD}$, $Acc^{i}_{STU}$ refer to the accuracies of SemCKD, a certain feature-map distillation and a regularly trained student model in the $i$-th setting, respectively. This evaluation metric reflects the extent to which SemCKD further improves on the basis of existing approaches compared to the improvement made by these approaches upon baseline student models. 

A computation-friendly metric called \textit{Semantic Mismatch score} (SM-score) is proposed as an alternative to the intractable nuclear norm calculation for $E^\prime$ in Equation (\ref{eq:nuclear}). SM-score is calculated by the average Euclidean distance between the generated similarity matrices of each associated teacher-student layer pair
\begin{equation}
	\label{eq:sms}
	\text{SM-score} = \frac{1}{|\mathcal{C}|} \sum_{\left(s_{l},t_{l}\right)\in \mathcal{C}}\bm{\alpha}_{(s_{l},t_{l})} \mathrm{MSE}\left(A^{s}_{s_{l}}, A^{t}_{t_{l}}\right),
\end{equation}
where $|\mathcal{C}|$ denotes the number of candidate layer pairs, $A^{s}_{s_{l}}$ and $A^{t}_{t_{l}}$ are similarity matrices of student layer $s_{l}$ and target layer $t_{l}$, respectively. 
$\bm{\alpha}_{(s_{l},t_{l})}$ always equal to $\bm{1}$ for compared approaches.
\subsubsection{Computing Infrastructure}
All of the experiments are implemented by PyTorch \cite{paszke2019pytorch} and conducted on a server containing one NVIDIA TITAN X-Pascal GPU as well as a server containing eight NVIDIA RTX 2080Ti GPUs. 

\begin{table}
	\caption{Top-1 test accuracy of  \textit{feature-map distillation} approaches on ImageNet. ResNet-34$^{*}$ denotes another ResNet-34 model trained with 10 more epochs.}
	\label{Table:ImageNet}
	\centering
	\begin{tabular}{cccc}
		\toprule
		\multirow{2}{*}{Student} & \small ResNet-18 & \small ShuffleV2x0.5 & \small ResNet-18 \\
		& 69.67 & 53.78 & 69.67 \\
		\midrule
		KD \cite{hinton2015distilling} & 70.62 & 53.73 & 70.54  \\
		FitNet \cite{romero2015fitnet} & 70.31 & 51.46 & 70.42  \\
		AT \cite{zagoruyko2017paying} & 70.30 &  52.83 & 70.30  \\
		SP \cite{tung2019similarity} & 69.99 & 51.73 & 70.12  \\
		VID \cite{ahn2019variational} & 70.30 & 53.97 & 70.26  \\
		HKD \cite{passalis2020heterogeneous} & 68.86 & 51.60 & 68.44  \\
		MGD \cite{yue2020mgd} & 70.37 & 52.96 & 70.18  \\
		\midrule
		SemCKD & \textbf{70.87} & \textbf{53.99} & \textbf{70.66}  \\
		\midrule
		\multirow{2}{*}{Teacher} & \small ResNet-34 & \small ResNet-34$^{*}$ & \small ResNet-34$^{*}$  \\
		& 73.26 & 73.54 &  73.54 \\
		\bottomrule
	\end{tabular}
\end{table}   

\subsection{Comparison of Different Feature-Map Distillation}
Table \ref{Table:CIFAR-100-1} and \ref{Table:CIFAR-100-2} give the Top-1 test accuracy (\%) on CIFAR-100 based on twelve network combinations, which consists of two homogeneous settings, i.e., the teacher and student share similar architectures (VGG-8/13, ResNet-8x4/32x4), and ten heterogeneous settings.
Each column apart from the first and last rows includes the results of a certain student model trained by various approaches under the supervision of the same teacher model. The results of the vanilla KD are also included for comparison. There are two combinations where the MGD \cite{yue2020mgd} is not applicable and denoted as '---'.

\begin{figure}
	\centering
	\includegraphics[width=0.9\columnwidth]{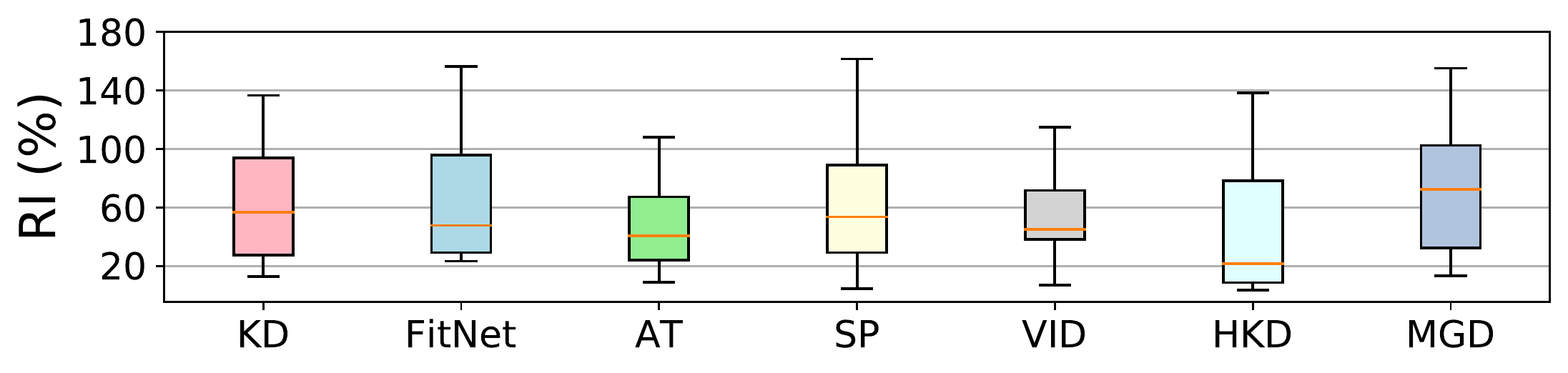}
	\caption{The distributions of \textit{Relative Improvement}.}
	\label{fig:ari}
\end{figure}

From Table \ref{Table:CIFAR-100-1} and \ref{Table:CIFAR-100-2}, we can see that SemCKD consistently achieves higher accuracy than state-of-the-art feature-map distillation approaches. 
The distributions of relative improvement are plotted in Figure \ref{fig:ari}. On average, SemCKD shows significantly relative improvement (60.03\%) over all these approaches. Specifically, the best case happens in the ``VGG-8 \& ResNet-32x4'' setting where the $RI$ over AT \cite{zagoruyko2017paying} is 242\%. When compared to the most competitive one named as HKD \cite{passalis2020heterogeneous}, which relies on a costly teacher-auxiliary-student paradigm as discussed in Section \ref{subsec:fmd}, the $RI$ becomes rather small on two settings (3.93\% for ``ShuffleNetV2 \& ResNet-32x4'', 9.98\% for ``MobileNetV2 \& WRN-40-2''). But in general, SemCKD still relatively outperforms HKD for 45.82\%, showing that our approach indeed make better use of intermediate information for effective distillation. 

We also find that none of the compared approaches can consistently beats the vanilla KD on CIFAR-100, which probably due to semantic mismatch among associated layer pairs. This problem becomes especially severe for one-pair selection (FitNet method fails in 7/12 cases and MGD method fails in 8/12 cases), the situation where the number of candidate layer $s_L$ is larger than $t_L$ (4/6 of methods fail in the ``ShuffleNetV2 \& VGG-13'' setting) and MobileNetV2 is used as the student model (5/6 or 4/6 of methods fail in the ``MobileNetV2 \& ResNet-32x4/WRN-40-2'' settings). Nevertheless, our semantic calibration formulation helps alleviate semantic mismatch to a great extent, leading to satisfied performance of SemCKD. Similar observations are obtained in Table \ref{Table:ImageNet} for the results on a large-scale image classification dataset.

\begin{figure*}
	\centering
	\includegraphics[width=0.98\textwidth]{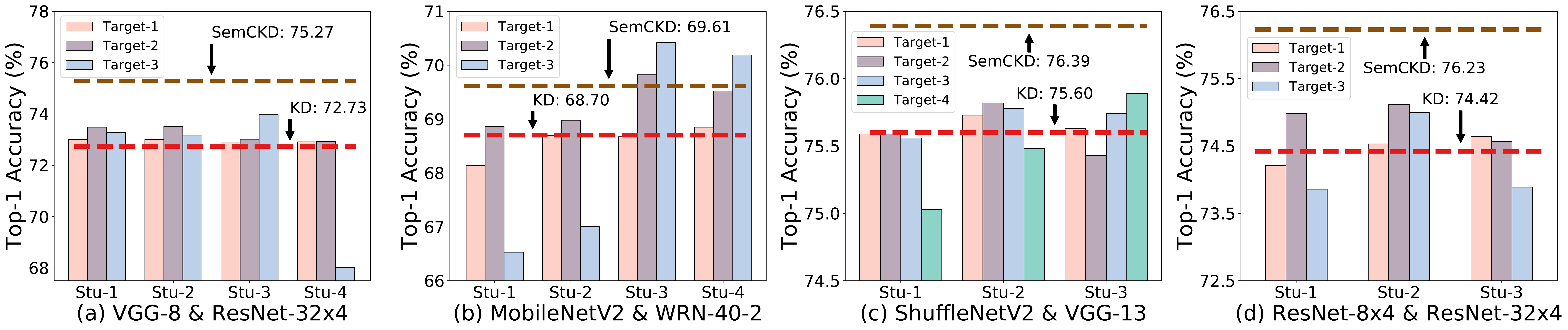}
	\caption{Illustration of negative regularization on CIFAR-100 with four model combinations. Each tick label of x-axis denotes a student layer number. The results for each specified teacher-student layer pair are shown with different color bars.}
	\label{fig:layer_match}
\end{figure*}

\subsection{Semantic Calibration Analysis}
\label{subsec:sem}
In this section, we experimentally study the negative regularization caused by manually specified layer associations and provide some explanations for the success of SemCKD by the proposed criterion and visual evidence.

\subsubsection{Negative regularization} 
\label{subsubsec:negative}
This effect occurs when feature-map distillation with a certain layer association performs poorer than the vanilla KD. To reveal its existence, we train the student model with only one specified teacher-student layer pair in various settings. The involved network architectures consist of ``VGG-8 \& ResNet-32x4'', ``MobileNetV2 \& WRN-40-2'', ``ShuffleNetV2 \& VGG-13'' and ``ResNet-8x4 \& ResNet-32x4''. The numbers of candidate target layers and student layers for each case are (3, 4), (3, 4), (4, 3) and (3, 3), respectively. 

Figure \ref{fig:layer_match} shows the results of student models with these 12 or 9 teacher-student layer combinations. For better comparison, results of the vanilla KD and SemCKD are plotted as dash horizontal lines with different colors. We can see that the performance of a student model becomes extremely poor for some layer associations, which is probably caused by large semantic gaps. Typical results include
``Student Layer-4 \& Target Layer-3'' in Figure \ref{fig:layer_match} (a), ``Student Layer-1, 2 \& Target Layer-3'' in Figure \ref{fig:layer_match} (b), ``Student Layer-1 \& Target Layer-4'' in Figure \ref{fig:layer_match} (c) and ``Student Layer-1, 3 \& Target Layer-3'' in Figure \ref{fig:layer_match} (d).

Another finding is that one-to-one layer association is suboptimal since better results can be achieved by exploiting information in a target layer with different depth, such as ``Student Layer-1 \& Target Layer-2'' in Figure \ref{fig:layer_match} (b), ``Student Layer-3 \& Target Layer-4'' in Figure \ref{fig:layer_match} (c) and ``Student Layer-1 \& Target Layer-2'' in Figure \ref{fig:layer_match} (d). Although training with certain hand-craft layer associations could outperform SemCKD in a few cases, such as ``Student Layer-3,4 \& Target Layer-3'' in Figure \ref{fig:layer_match} (b), SemCKD still performs reasonably well against a large selection of associations, especially knowledge of the best layer association for any of network combinations is not available in advance. Nevertheless, those cases in which SemCKD is inferior to the best one indicate that there is extra room for refinement of our association strategy.

\begin{table}
	\caption{Evaluation of semantic match for VGG-8 \& ResNet-32x4 on CIFAR-100.}
	\label{Table:semantic}
	\centering
	\begin{tabular}{ccc}
		\toprule
		Approach & SM-score $(\downarrow)$ & CKA $(\uparrow)$ \\ 
		\midrule
		FitNet \cite{romero2015fitnet} & 13.38 & \textbf{0.99} \\
		AT \cite{zagoruyko2017paying} & 15.52 & 0.90 \\
		SP \cite{tung2019similarity} & 16.30 & 0.91 \\
		VID \cite{ahn2019variational} & 15.82 & 0.91 \\
		HKD \cite{passalis2020heterogeneous} & 19.03 & 0.95 \\ 
		MGD \cite{yue2020mgd} & 17.80 & 0.70 \\
		\midrule
		SemCKD & \textbf{11.27} & \textbf{0.98} \\
		\bottomrule
	\end{tabular}
\end{table}
\subsubsection{Semantic Mismatch Score}
We then evaluate whether SemCKD actually leads to less semantic mismatch solutions compared with other approaches. A new metric called SM-score is proposed as Equation (\ref{eq:sms}), which hopefully represents the degree of difference between the captured pairwise similarity among instances in certain semantic level. Additionally, we adopt CKA \cite{kornblith2019similarity} to evaluate the similarity between extracted feature maps as the previous work \cite{xu2020knowledge}. We calculate the SM-scores (log-scale) and CKA values of each approach across training epochs and average them in the last 10 epochs, which already keep almost unchanged, as the final ones reported in Table \ref{Table:semantic}. $\uparrow (\downarrow)$ in the Table \ref{Table:semantic} indicates the larger (smaller) the better. Thanks to our soft layer association, the lowest semantic mismatch score and nearly the largest CKA through the training process is achieved by SemCKD.

\subsubsection{Visualization}
To further provide visual explanations for the advantage of SemCKD, we randomly select several images from ImageNet labeled by ``Bow tie'', ``Rain barrel'', ``Racer'', ``Bathhub'', ``Goose'', ``Band aid'', ``Sweatshirt'', ``Saltshaker'', ``Rock crab'' and ``Woollen'', and use Grad-CAM \cite{Selvaraju17grad} to highlight the regions which are considered to be important for a model to predict the corresponding labels.

\begin{figure*}
	\centering
	\includegraphics[width=0.99\textwidth]{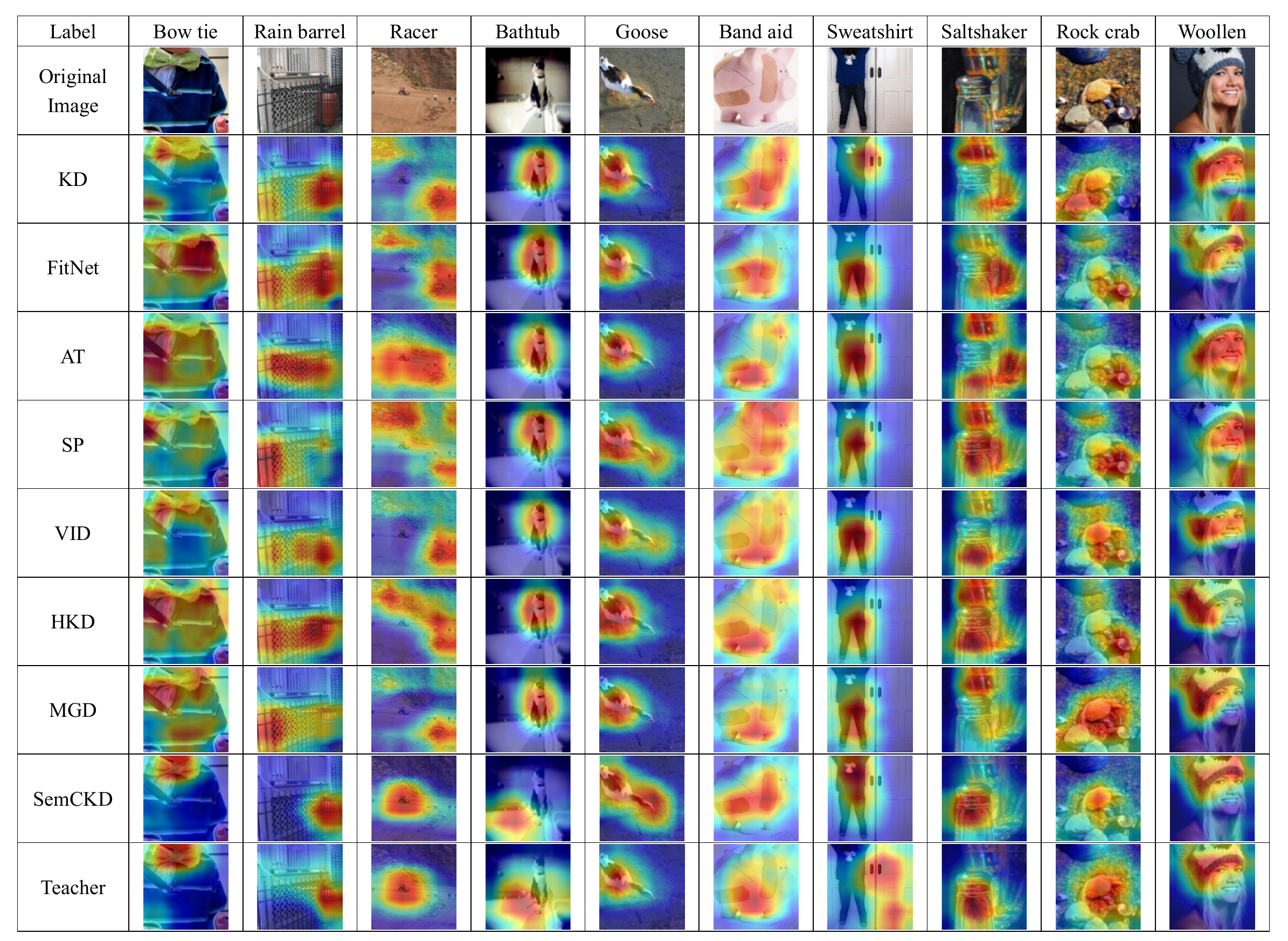}
	\caption{Grad-CAM visualization of \textit{feature-map distillation} approaches on ImageNet. The first two rows include the original images and their associated labels. The third to the tenth rows show the visualization results of each approach. Red region is more important for the model prediction. Figure best viewed in color.}
	\label{fig:vis}
\end{figure*}

From Figure \ref{fig:vis}, we can see that the class-discriminative regions are centralized by SemCKD, which is similar to the teacher model in most cases, while being scatted around the surroundings by compared approaches. As visualized in the fifth column, where the images are labeled with ``Bathtub", another failure mode of the compared approaches is that they sometimes regard the right regions as background while putting their attention on the spatial adjacency object. As for the images in the eighth column labeled with ``Sweatshirt", the teacher model puts its attention on a wrong region corresponding to the label ``sliding door", which leads the student model trained with vanilla KD to make the same mistake and leads other compared approaches to focus on the wrong region corresponding to the label ``jean''. 
Moreover, SemCKD can capture more semantic-related information like highlighting the head and neck to identify a ``Goose'' in the image. 

\begin{table*}
	\caption{Ablation study: Top-1 test accuracy for VGG-8 \& ResNet-32x4 on CIFAR-100.}
	\label{Table-ab}
	\centering
	\begin{tabular}{cccccc}
		\toprule
		Equal Allocation & w/o Projection & w/o $\mathrm{MLP}$ & w/o Similarity Matrices &  Shared Allocation  & SemCKD  \\
		\midrule
		72.94 $\pm$ 0.87 & 72.51 $\pm$ 0.16 & 72.78 $\pm$ 0.29 & 74.75 $\pm$ 0.14 & 74.87 $\pm$ 0.21 & 75.27 $\pm$ 0.13 \\
		\bottomrule
	\end{tabular}
\end{table*}

\begin{figure*}
	\centering
	\includegraphics[width=0.98\textwidth]{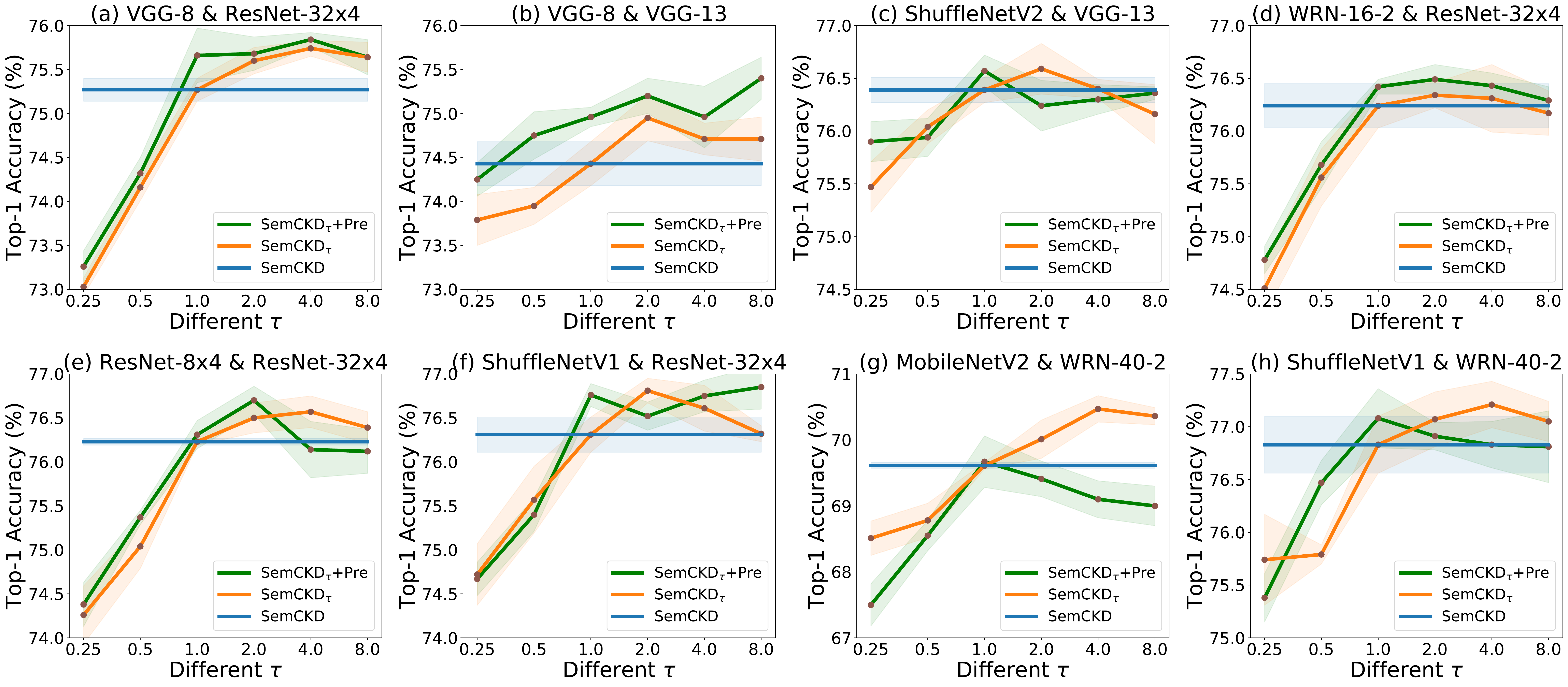}
	\caption{Illustration of the effect of softening attention on CIFAR-100 with eight different model combinations.}
	\label{fig:tau}
\end{figure*}
\begin{figure*}
	\centering
	\includegraphics[width=0.98\textwidth]{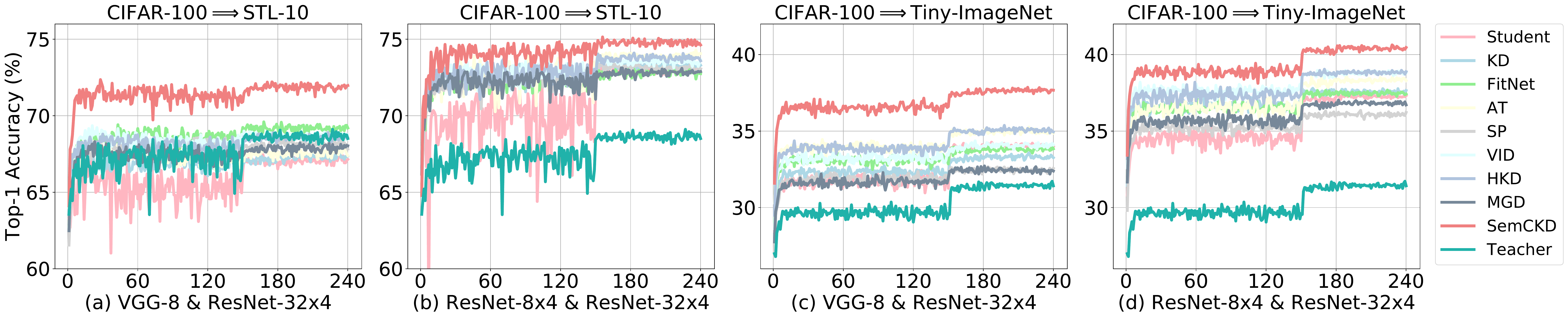}
	\caption{Transfer Learning scenario: Top-1 test accuracy for two combinations from CIFAR-100 to STL-10 or Tiny-ImageNet.}
	\label{fig:transfer}
\end{figure*} 

\subsection{Ablation Study} 
\label{subsec:ab}
In Table \ref{Table-ab}, we present the evaluation results of five kinds of ablation studies based on the ``VGG-8 \& ResNet-32x4'' combination.
Specifically, the first kind confirms the benefit of adaptively learning attention allocation for cross-layer knowledge distillation, the second to the fourth ones confirm the benefit of each individual component for obtaining attention weights, and the last one confirms the benefit of assigning each instance with independent weights.

(1) Equal Allocation. In order to validate the effectiveness of allocating the personalized attention of each student layer to multiple target layers, equal weights assignment is applied instead. This causes a lower accuracy by 2.33\% (from 75.27\% to 72.94\%) and a considerably larger variance by 0.74\% (from 0.13\% to 0.87\%).

(2) w/o Projection. Rather than projecting the feature maps of each student layer to the same dimension as those in the target layers by Equation (\ref{eq:proj}), we add a new $\mathrm{MLP}_V(\cdot)$ to project the pairwise similarity matrices of teacher-student layer pairs into another subspace to generate \textit{value} vectors. Thus the Mean-Square-Error among feature maps in Equation (\ref{eq:semCKD}) is replaced by these \textit{value} vectors to calculate the overall loss, which reduces the performance by 2.76\%.

(3) w/o $\mathrm{MLP}$. A simple linear transformation is used to obtain \textit{query} and \textit{key} vectors in Equation (\ref{eq:mlp}) instead of the two-layer non-linear transformation $\mathrm{MLP}(\cdot)$, which includes ``Linear-ReLU-Linear-Normalization''.  
The 2.49\% performance drop indicates that the usefulness of $\mathrm{MLP}(\cdot)$ to alleviate the effect of noise and sparseness.

(4) w/o Similarity Matrices. We skip the similarity matrices calculation in Equation (\ref{eq:adj}) and directly employ feature maps of candidate layer pairs to obtain the \textit{query} and \textit{key} vectors. Specifically, we replace Equation (\ref{eq:mlp}) with 
$Q_{s_{l}}[i]=\mathrm{MLP}_{Q}(F^{s}_{s_{l}}[i])$, $ 
K_{t_{l}}[i]=\mathrm{MLP}_{K}(F^{t}_{t_{l}}[i]).$
The 0.52\% performance decline indicates the benefit of similarity matrices for the subsequent attention calculation. Another reason for not using feature maps is that this will cause a considerable memory cost to generate \textit{query} and \textit{key} vectors due to the larger spatial dimensions ($c_{s_l/t_l}\cdot h_{s_l/t_l}\cdot w_{s_l/t_l}\gg b$). 

(5) Shared Allocation. Instead of learning an independent attention allocation across teacher-student layer pairs for each training instance, we make all instances share the same association weights during the whole training process.
This will incur 0.40\% accuracy drop.

\subsection{Softening Attention}
\label{subsec:softening}
In this section, we conduct experiments on a large number of teacher-student combinations to verify the effectiveness of softening attention in SemCKD. 

\begin{table*}[htbp]
	\centering
	\caption{Few-Shot Learning scenario: Top-1 test accuracy for VGG-8 \& ResNet-32x4 on CIFAR-100.}
	\label{Table:few-shot}
	\resizebox{0.64\textwidth}{!}{
		\begin{tabular}{c|cccc}
			\toprule
			Percentage & 25\% & 50\% & 75\% & 100\% \\
			\midrule
			Student & 55.26 $\pm$ 0.33 & 64.28 $\pm$ 0.20 & 68.21 $\pm$ 0.16 & 70.46 $\pm$ 0.29 \\
			\midrule
			KD \cite{hinton2015distilling} & 59.23 $\pm$ 0.23 & 67.16 $\pm$ 0.14 & 70.32 $\pm$ 0.16 & 72.73 $\pm$ 0.15 \\
			FitNet \cite{romero2015fitnet} & 60.29 $\pm$ 0.32 & 67.58 $\pm$ 0.25 & 70.84 $\pm$ 0.33 & 72.91 $\pm$ 0.18 \\
			AT \cite{zagoruyko2017paying} & 59.72 $\pm$ 0.21 & 66.94 $\pm$ 0.13 & 70.18 $\pm$ 0.36 & 71.90 $\pm$ 0.13 \\
			SP \cite{tung2019similarity} & 60.93 $\pm$ 0.22 & 67.92 $\pm$ 0.22 & 71.22 $\pm$ 0.28 & 73.12 $\pm$ 0.10 \\
			VID \cite{ahn2019variational} & 60.15 $\pm$ 0.15 & 67.91 $\pm$ 0.19 & 70.86 $\pm$ 0.15 & 73.19 $\pm$ 0.23 \\
			HKD \cite{passalis2020heterogeneous} & 58.96 $\pm$ 0.24 & 67.25 $\pm$ 0.31 & 70.38 $\pm$ 0.26 & 72.63 $\pm$ 0.12 \\
			MGD \cite{yue2020mgd} & 60.03 $\pm$ 0.17 & 67.31 $\pm$ 0.14 & 70.59 $\pm$ 0.29 & 72.39  $\pm$ 0.16\\
			\midrule
			SemCKD & \textbf{64.82 $\pm$ 0.20} & \textbf{70.76 $\pm$ 0.23} & \textbf{73.06 $\pm$ 0.25} & \textbf{75.27 $\pm$ 0.13}  \\
			\bottomrule
		\end{tabular}
	}
\end{table*}   
\begin{table*}[htbp]
	\centering
	\caption{Nosiy-Label Learning scenario: Top-1 test accuracy for VGG-8 \& ResNet-32x4 on CIFAR-100.}
	\label{Table:nosiy}
	\resizebox{0.9\textwidth}{!}{
		\begin{tabular}{c|cccccc}
			\toprule
			Percentage & 0\% & 10\% & 20\% & 30\% & 40\% & 50\% \\
			\midrule
			Student & 70.46 $\pm$ 0.29 & 66.48 $\pm$ 0.27 & 60.08 $\pm$ 0.16 & 49.57 $\pm$ 0.11 & 39.93 $\pm$ 0.15 & 32.81 $\pm$ 0.11 \\
			\midrule
			KD \cite{hinton2015distilling} & 72.73 $\pm$ 0.15 & 71.39 $\pm$ 0.33 & 67.99 $\pm$ 0.20 & 62.53 $\pm$ 0.29 & 58.70 $\pm$ 0.21 & 54.54 $\pm$ 0.13 \\
			FitNet \cite{romero2015fitnet} & 72.91 $\pm$ 0.18 & 71.59 $\pm$ 0.23 & 68.25 $\pm$ 0.20 & 62.68 $\pm$ 0.17 & 58.94 $\pm$ 0.39 & 54.93 $\pm$ 0.19 \\
			AT \cite{zagoruyko2017paying} & 71.90 $\pm$ 0.13 & 70.81 $\pm$ 0.19 & 67.75 $\pm$ 0.29 & 61.81 $\pm$ 0.15 & 57.94 $\pm$ 0.27 & 54.08 $\pm$ 0.30 \\
			SP \cite{tung2019similarity} & 73.12 $\pm$ 0.10 & 71.73 $\pm$ 0.27 & 68.49 $\pm$ 0.18 & 62.94 $\pm$ 0.38 & 59.21 $\pm$ 0.29 & 54.81 $\pm$ 0.39 \\
			VID \cite{ahn2019variational} & 73.19 $\pm$ 0.23 & 72.02 $\pm$ 0.08 & 68.72 $\pm$ 0.14 & 63.22 $\pm$ 0.27 & 59.26 $\pm$ 0.40 & 55.26 $\pm$ 0.29 \\
			HKD \cite{passalis2020heterogeneous} & 72.63 $\pm$ 0.12 & 68.87 $\pm$ 0.31 & 62.71 $\pm$ 0.14 & 52.41 $\pm$ 0.14 & 43.84 $\pm$ 0.06 & 36.94 $\pm$ 0.18 \\
			MGD \cite{yue2020mgd} & 72.39 $\pm$ 0.16 & 71.34 $\pm$ 0.08 & 68.09 $\pm$ 0.19 & 62.56 $\pm$ 0.27 & 58.65 $\pm$ 0.31 & 54.57 $\pm$ 0.39 \\
			\midrule
			SemCKD & \textbf{75.27 $\pm$ 0.13} & \textbf{73.81 $\pm$ 0.19} & \textbf{70.62 $\pm$ 0.15} & \textbf{65.16 $\pm$ 0.10} & \textbf{60.93 $\pm$ 0.06} & \textbf{57.19 $\pm$ 0.19} \\
			\bottomrule
		\end{tabular}
	}
\end{table*}   
As shown in Figure \ref{fig:tau}, we test the performance of each student model with different softness $\tau$ and draw an orange curve for the results of SemCKD$_{\tau}$. To preserve as much information as possible in the teacher model during knowledge transfer, previous works attempt to adjust the distillation position to the front of the ReLU operation
\cite{heo2019fd,yue2020mgd}, which is called \textit{pre-activation} distillation. We also combine this operation into SemCKD$_{\tau}$ and name this variant as SemCKD$_{\tau}$+Pre. Additionally, the results of original SemCKD are plotted as the horizontal lines for comparison.

In most cases, we can see that softening attention indeed improves the performance of SemCKD by a considerable margin. For example, when $\tau=4$, the absolute accuracy boost is 0.47\% and 0.86\% in the ``VGG-8 \& ResNet-32x4'' and ``MobileNetV2 \& WRN-40-2'' settings, respectively. Another evidence to support the necessity of softness is that there will be a significant performance degeneration when $\tau$ is less than 1. The reason is that the attention weights will be sharpened in this situation and make the overall target direction become largely influenced by a certain component. Generally, $\tau=2$ or $4$ is a satisfying initial choice in practice for a given teacher-student combination. We also find that SemCKD$_{\tau}$+Pre could only outperform SemCKD$_{\tau}$ slightly but will be inferior to SemCKD distinctly when ``MobileNetV2'' is used for the student model. 

\subsection{Generalization to Different Scenarios}
\subsubsection{Transfer Learning}

We conduct a series of experiments on two dataset pairs with two network combinations to evaluate the transferability of learned representations.
Each model trained with the corresponding approach on CIFAR-100 is adopted to extract representations for the images from STL-10 or Tiny-ImageNet. Then, we train another linear classifier on these representations for evaluation.

As shown in Figure \ref{fig:transfer}, SemCKD beats all compared approaches by a considerable margin. Specifically, SemCKD outperforms the second best one (FitNet) by 2.76\% absolute accuracy in Figure \ref{fig:transfer} (a). Although representations learned by the teacher model transfer worst in the most settings (3/4) as discussed in \cite{tian2020contrastive}, SemCKD still absolutely improves the performance of student models by 3.25\% on average.

\subsubsection{Few-Shot Learning and Noisy-Label Learning}
We further evaluate the performance of SemCKD in few-shot learning and noisy-label learning scenarios. All these experiments indicate that SemCKD can make better use of training data and is relatively robust to noisy perturbation.

Following the setting of \cite{xu2020knowledge}, we randomly sample 25\%, 50\% and 75\% training images from each class on CIFAR-100 to make several few-shot learning datasets, and randomly perturb 10\%, 20\%, 30\%, 40\% and 50\% labels of training images to make noisy-label learning datasets.
From Table \ref{Table:few-shot} and \ref{Table:nosiy}, we can see that SemCKD consistently outperforms compared approaches as well as the regular training in all settings. The improvement will become larger as the number of available training images reduces or the number of perturbed training images increases. Specifically, as shown in Table \ref{Table:few-shot}, SemCKD can surpass the regularly trained student model by training with only 50\% original dataset and become comparable to state-of-the-art approaches given additional 25\% training images. 
Similarly, from Table \ref{Table:nosiy}, we can see that SemCKD can surpass state-of-the-art approaches by training with 10\% noisy data, and surpass the regularly trained student model even with additional 10\% noisy data in training images. 

\begin{table*}[htbp]
	\centering
	\caption{Top-1 test accuracy of \textit{feature-embedding distillation} approaches on CIFAR-100.}
	\label{Table:FE-CIFAR-100}
	\begin{tabular}{c|ccccccc|c}
		\toprule
		Student & PKT \cite{passalis2018learning} & RKD  \cite{park2019relational} & IRG \cite{liu2019knowledge} & CC \cite{peng2019correlation} & CRD \cite{tian2020contrastive} & SemCKD & SemCKD+CRD  & Teacher  \\
		\midrule
		VGG-8 & \multirow{2}{*}{73.11 $\pm$ 0.21} & \multirow{2}{*}{72.49 $\pm$ 0.08} & \multirow{2}{*}{72.57 $\pm$ 0.20} &  \multirow{2}{*}{72.63 $\pm$ 0.30} & \multirow{2}{*}{73.54 $\pm$ 0.19} & 
		\multirow{2}{*}{75.27 $\pm$ 0.13} & \multirow{2}{*}{\textbf{75.52 $\pm$ 0.09}} &
		ResNet-32x4 \\
		70.46 $\pm$ 0.29 & & & & & & & & 79.42 \\
		\midrule
		VGG-13 & 
		\multirow{2}{*}{77.43 $\pm$ 0.11} & \multirow{2}{*}{76.93 $\pm$ 0.19} &
		\multirow{2}{*}{76.91 $\pm$ 0.15} & \multirow{2}{*}{77.16 $\pm$ 0.19} & \multirow{2}{*}{77.48 $\pm$ 0.13} & \multirow{2}{*}{79.43 $\pm$ 0.02} & \multirow{2}{*}{\textbf{79.48 $\pm$ 0.07}} & ResNet-32x4 \\
		74.82 $\pm$ 0.22 & & & & & & & & 79.42 \\	
		\midrule
		ShuffleNetV2 & \multirow{2}{*}{75.82 $\pm$ 0.28} & 
		\multirow{2}{*}{75.51 $\pm$ 0.34} & \multirow{2}{*}{75.65 $\pm$ 0.39} & \multirow{2}{*}{75.40 $\pm$ 0.19} & \multirow{2}{*}{76.11 $\pm$ 0.27} & \multirow{2}{*}{76.39 $\pm$ 0.12} & \multirow{2}{*}{\textbf{76.58 $\pm$ 0.29}} &
		VGG-13  \\
		72.60 $\pm$ 0.12 & & & & & & & & 74.64 \\
		\midrule
		ShuffleNetV2 & \multirow{2}{*}{75.88 $\pm$ 0.06} & \multirow{2}{*}{75.67 $\pm$ 0.25} & \multirow{2}{*}{75.45 $\pm$ 0.27} & \multirow{2}{*}{75.47 $\pm$ 0.28} & \multirow{2}{*}{77.04 $\pm$ 0.61} & \multirow{2}{*}{77.62 $\pm$ 0.32} & \multirow{2}{*}{\textbf{77.94 $\pm$ 0.33}} & ResNet-32x4 \\
		72.60 $\pm$ 0.12 & & & & & & & & 79.42 \\
		\midrule
		MobileNetV2 & 
		\multirow{2}{*}{68.68 $\pm$ 0.29} & \multirow{2}{*}{68.71 $\pm$ 0.20} &
		\multirow{2}{*}{68.83 $\pm$ 0.18} & \multirow{2}{*}{68.68 $\pm$ 0.14} & \multirow{2}{*}{69.98 $\pm$ 0.27} & \multirow{2}{*}{69.61 $\pm$ 0.05} & \multirow{2}{*}{\textbf{70.55 $\pm$ 0.11}} & WRN-40-2 \\
		65.43 $\pm$ 0.29 & & & & & & & & 75.61 \\
		\midrule
		ShuffleNetV1 &  \multirow{2}{*}{74.48 $\pm$ 0.16} & \multirow{2}{*}{74.22 $\pm$ 0.26} & \multirow{2}{*}{74.11 $\pm$ 0.12} & \multirow{2}{*}{74.26 $\pm$ 0.49} & \multirow{2}{*}{75.34 $\pm$ 0.24} & \multirow{2}{*}{76.31 $\pm$ 0.20} & \multirow{2}{*}{\textbf{76.60 $\pm$ 0.08}} & ResNet-32x4 \\
		71.36 $\pm$ 0.25 && & & & & & & 79.42\\
		\midrule
		WRN-16-2 &  \multirow{2}{*}{75.07 $\pm$ 0.12} & \multirow{2}{*}{74.93 $\pm$ 0.45} & \multirow{2}{*}{74.73 $\pm$ 0.23} & \multirow{2}{*}{74.91 $\pm$ 0.15} & \multirow{2}{*}{75.89 $\pm$ 0.38} & \multirow{2}{*}{76.24 $\pm$ 0.21} & \multirow{2}{*}{\textbf{76.32 $\pm$ 0.14}} & ResNet-32x4 \\
		73.41 $\pm$ 0.27   && & & & & &  & 79.42 \\
		\midrule
		ShuffleNetV1 & \multirow{2}{*}{75.42 $\pm$ 0.24} & \multirow{2}{*}{75.25 $\pm$ 0.16} & \multirow{2}{*}{75.24 $\pm$ 0.23} & \multirow{2}{*}{75.23 $\pm$ 0.22} & \multirow{2}{*}{75.89 $\pm$ 0.24} & \multirow{2}{*}{76.83 $\pm$ 0.27} & \multirow{2}{*}{\textbf{76.92 $\pm$ 0.26}}  & WRN-40-2 \\
		71.36 $\pm$ 0.25 && & & & & & & 75.61 \\
		\midrule
		MobileNetV2 & 
		\multirow{2}{*}{67.99 $\pm$ 0.29} & \multirow{2}{*}{67.46 $\pm$ 0.55} &
		\multirow{2}{*}{67.89 $\pm$ 0.26} & \multirow{2}{*}{67.68 $\pm$ 0.22} & \multirow{2}{*}{69.39 $\pm$ 0.11} & \multirow{2}{*}{69.00 $\pm$ 0.19} & \multirow{2}{*}{\textbf{69.81 $\pm$ 0.42}} & ResNet-32x4 \\
		65.43 $\pm$ 0.29 & & & & & & & & 79.42 \\
		\midrule
		MobileNetV2 & 
		\multirow{2}{*}{69.57 $\pm$ 0.14} & \multirow{2}{*}{69.39 $\pm$ 0.41} &
		\multirow{2}{*}{69.51 $\pm$ 0.31} & \multirow{2}{*}{70.00 $\pm$ 0.17} & \multirow{2}{*}{70.83 $\pm$ 0.17} & \multirow{2}{*}{70.27 $\pm$ 0.12} & \multirow{2}{*}{\textbf{71.07 $\pm$ 0.07}} & ShuffleNetV2 \\
		65.43 $\pm$ 0.09 & & & & & & & & 72.60 \\
		\midrule
		VGG-8 & \multirow{2}{*}{73.40 $\pm$ 0.30} & \multirow{2}{*}{73.42 $\pm$ 0.14} & \multirow{2}{*}{73.31 $\pm$ 0.12} & \multirow{2}{*}{73.33 $\pm$ 0.23} & \multirow{2}{*}{74.31 $\pm$ 0.17} & \multirow{2}{*}{74.43 $\pm$ 0.25} & \multirow{2}{*}{\textbf{74.67 $\pm$ 0.11}} & VGG-13 \\
		70.46 $\pm$ 0.29 & & & & & & & & 74.64 \\
		\midrule
		ResNet-8x4& 
		\multirow{2}{*}{74.61 $\pm$ 0.25} & \multirow{2}{*}{74.36 $\pm$ 0.23} &
		\multirow{2}{*}{74.67 $\pm$ 0.15} & \multirow{2}{*}{74.50 $\pm$ 0.13} & \multirow{2}{*}{75.59 $\pm$ 0.07} & \multirow{2}{*}{76.23 $\pm$ 0.04} & \multirow{2}{*}{\textbf{76.68 $\pm$ 0.19}} & ResNet-32x4 \\
		73.09 $\pm$ 0.30 & & & & & & & & 79.42 \\
		\bottomrule   
	\end{tabular}
\end{table*}

\subsection{Extension to Feature-Embedding Distillation}
\label{subsec:fed}
Knowledge transfer based on feature embeddings of the penultimate layer is another alternative to improve the generalization ability of student models. We compare the performance of several newly proposed approaches on the same twelve network combinations as Table \ref{Table:CIFAR-100-1} and Table \ref{Table:CIFAR-100-2}. Additionally, we report the results of student models trained by simply adding the loss of CRD \cite{tian2020contrastive} into the original one of SemCKD without tuning any hyper-parameter. 

From Table \ref{Table:FE-CIFAR-100}, we can observe that the performance can be further boosted by training with ``SemCKD+CRD'' in all cases, which confirms that our approach holds a very satisfying property to be highly compatible with the state-of-the-art feature-embedding distillation approach. 

\subsection{Sensitivity Analysis} 
\begin{figure}
	\centering
	\includegraphics[width=0.96\columnwidth]{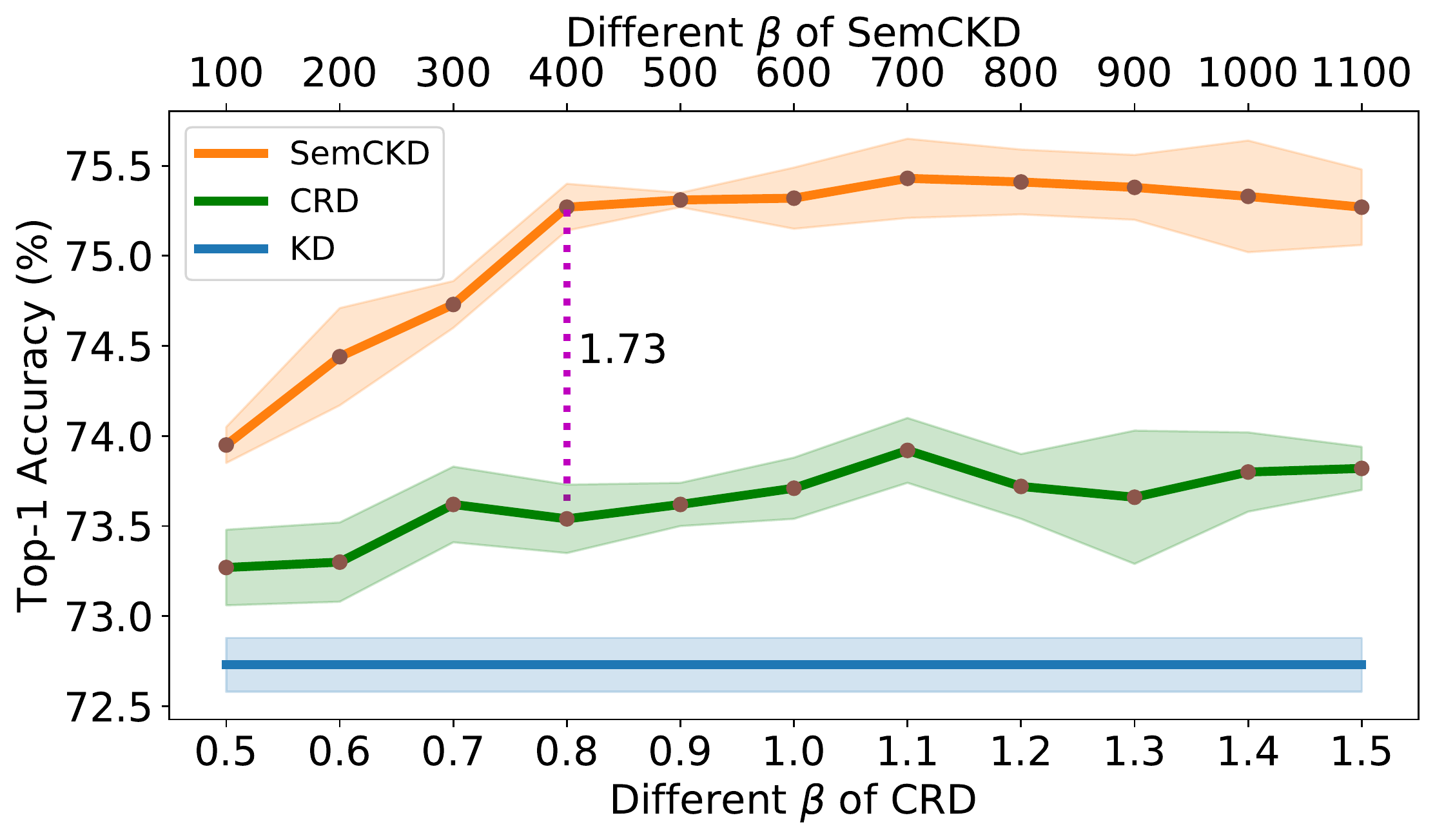}
	\caption{Impact of the hyper-parameter $\beta$ for VGG-8 \& ResNet-32x4 on CIFAR-100.}
	\label{fig:Beta}
\end{figure}

Finally, we evaluate the impact of hyper-parameter $\beta$ on the performance. 
We compare three representative knowledge distillation approaches, including logits distillation (KD \cite{hinton2015distilling}), feature-embedding distillation (CRD \cite{tian2020contrastive}) and feature-map distillation (SemCKD). The range of hyper-parameter $\beta$ for SemCKD is set as 100 to 1100 at equal interval of 100, while the hyper-parameter $\beta$ for CRD ranges from 0.5 to 1.5 at equal interval of 0.1, adopting the same search space as the original paper \cite{tian2020contrastive}. Note that the hyper-parameter $\beta$ always equals to $0$ for the vanilla KD, leading to a horizontal line in Figure \ref{fig:Beta}.

It is seen that SemCKD achieves the best results in all cases and outperforms CRD at about 1.73\% absolute accuracy for the default hyper-parameter setting. 
Figure \ref{fig:Beta} also shows that the performance of SemCKD keeps very stable after the hyper-parameter $\beta$ is greater than 400, which indicates that our proposed method works reasonably well in a wide range of search space for the hyper-parameter $\beta$. 

\section{Conclusion}

In this paper, we focus on a critical but neglected issue in feature-map based knowledge distillation, i.e., how to alleviate performance degeneration resulted from negative regularization in manually specified layer pairs.
Our strategy is to use an attention mechanism for association weight learning, based on which knowledge could be transferred in a matched semantic space.
We further provide empirical evidence to support the semantic calibration capability of SemCKD and connect the association weights with the classic Orthogonal Procrustes problem. Extensive experiments show that SemCKD consistently outperforms the compared state-of-the-art approaches and our softening attention variant SemCKD$_{\tau}$ further widens the lead.
Moreover, our approach is readily applicable to different tasks, network architectures, and highly compatible with the feature-embedding distillation approach.

\ifCLASSOPTIONcaptionsoff
\newpage
\fi


\bibliographystyle{IEEEtran}
\bibliography{ref_KD.bib}
%




\end{document}